\pdfoutput=1 
\documentclass{article}

\usepackage[preprint]{neurips_2026}
\PassOptionsToPackage{expansion=false}{microtype}
\usepackage{microtype}
\usepackage{graphicx}
\usepackage{subcaption}
\usepackage{booktabs}
\usepackage{multirow}
\usepackage{float}

\usepackage{amsmath}
\usepackage{amssymb}
\usepackage{mathtools}
\usepackage{amsthm}

\usepackage[utf8]{inputenc}
\usepackage[T1]{fontenc}
\usepackage{algorithm}
\usepackage{algpseudocode}
\usepackage{hyperref}
\hypersetup{
  pdftitle={Structurally Separated DAG Learning with Multi-Scale Normalized Closure},
  pdfauthor={Shuai Feng, Hao Yan}
}
\usepackage{url}
\usepackage{amsfonts}
\usepackage{nicefrac}
\usepackage{xcolor}

\theoremstyle{plain}
\newtheorem{theorem}{Theorem}[section]
\newtheorem{proposition}[theorem]{Proposition}
\newtheorem{lemma}[theorem]{Lemma}
\newtheorem{corollary}[theorem]{Corollary}
\theoremstyle{definition}

\theoremstyle{remark}
\newtheorem{remark}[theorem]{Remark}

\let\cite\citep

\graphicspath{{./}}

\title{Structurally Separated DAG Learning with Multi-Scale Normalized Closure}

\author{%
  Shuai Feng\\
  Arizona State University\\
  \texttt{sfeng51@asu.edu}
  \And
  Hao Yan\\
  Arizona State University\\
  \texttt{hyan46@asu.edu}
}

\begin{document}

\maketitle

\begin{abstract}
Continuous DAG learning commonly enforces acyclicity through scalar functions of a weighted adjacency that aggregate evidence from weighted closed walks.
We add a complementary layer: matrix-valued closure logits---explicit multi-hop structural relation scores for each ordered pair---supervised alongside data fit, sparsity, and a DAGMA-style barrier on the weighted adjacency.
The distinction is not scalarization of the overall objective but how path evidence is represented and exposed before aggregation.
\emph{SMNC} (scale-normalized multi-scale closure) builds these scores on a structural gate by repeated squaring in log space and divides each scale by its path-length budget, counteracting product-style attenuation along long cycles; diagonal entries support node-wise cycle supervision and off-diagonals optional pairwise reachability cues.
In the linear model, a bounded structure--magnitude factorization ties strong effective edges to strong gates; a nonlinear induced gate offers another construction with the same intent.
For sufficiently strong cycles, normalized logits admit a length-independent lower bound; under a unique-active-cycle assumption, the hard final-scale readout yields a polynomial log-gate gradient factor rather than exponential product-style decay (Section~\ref{sec:smnc-properties}).
On cue-free, unstandardized Gaussian linear and nonlinear ER benchmarks, these models improve over DAGMA, run with default hyperparameters at a reduced budget (Appendix~\ref{app:baseline-protocol}), on denser graphs and are close to it on sparse ones.
\end{abstract}

\section{Introduction}

Learning directed acyclic graphs (DAGs) from observational data is a central problem in causal discovery and graphical modeling. Exact score-based search over DAGs is NP-hard~\cite{nphard,chickering2004large}, motivating both classical combinatorial families and continuous relaxations that optimize weighted adjacencies with gradient methods~\cite{spirtes2000causation,chickering2002optimal,heckerman1995learning,tsamardinos2006maxmin,gasse2014hybrid,raskutti2018sparsest,solus2021greedy,heinze2018causal,kitson2023survey,notears,nofears,changswap,dagma}. Recent analyses also emphasize that practical behavior depends strongly on optimization geometry, identifiability assumptions, and benchmark design~\cite{ng2022convergence,deng2023global,deng2024markov,ng2024sober,reisach2021beware}.

Continuous DAG learning commonly enforces acyclicity through scalar functions of a weighted adjacency. These objectives aggregate evidence from weighted closed walks, which can exhibit strong dependence on path length, edge magnitude, and repeated cycle traversals. We investigate a complementary representation: a matrix-valued closure whose entries provide explicit multi-hop structural relation scores. This representation permits node-wise cycle supervision and pairwise reachability constraints, while scale normalization mitigates attenuation along long paths. Our full objective still combines data fit, sparsity, a DAGMA-style log-determinant barrier on the weighted adjacency, and closure-side losses; what distinguishes SMNC is how path evidence is represented and exposed before this scalar aggregation, not the presence of a scalar barrier itself. A graph is acyclic exactly when strict diagonal self-reachability is zero, so cycle evidence can be read out per node; off-diagonal logits can absorb domain relations such as ancestry~\cite{chen2016ancestral,li2018sideconstraints,differentialorder,ordereffective_4} that are awkward to state as one-hop edge labels. This does not imply that DAGMA lacks a closure or that its scalar barrier is inherently defective. Section~\ref{sec:related} reviews walk-based acyclicity penalties in this light.

\begin{figure}[t]
  \centering
  \includegraphics[width=0.72\linewidth]{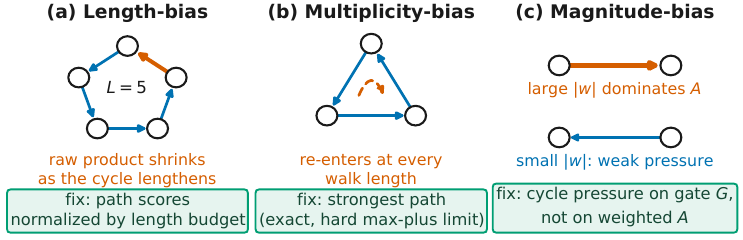}
  \caption[Effects of pooling weighted walks into a scalar acyclicity penalty, and the corresponding closure-head design choices]{Effects that can arise when weighted-walk evidence is pooled into scalar acyclicity objectives, and corresponding design choices in our matrix-valued closure head: scale-normalized path scores mitigate length-dependent attenuation, the hard max-plus strongest-path limit reduces multiplicity from repeated traversals, and supervising the gate $G$ rather than coefficient magnitudes in $A=W_f(M)\odot G(Z)$ separates structural support from effect size.}
  \label{fig:bias-overview}
\end{figure}

When weighted-walk evidence enters only through pooled acyclicity terms, optimization can couple strongly to walk length, multiplicity, and edge magnitude (Figure~\ref{fig:bias-overview}; Appendix~\ref{app:walk_accounting}). NOTEARS-style and DAGMA objectives exemplify this pattern~\cite{notears,nofears,dagma,changswap}: matrix powers or resolvent entries encode walks, but a single scalar (trace, log-determinant, or spectral quantity) aggregates them. Familiar effects include \textbf{length dependence} through products along long cycles and coefficient schedules; \textbf{multiplicity} when the same simple cycle is counted at every multiple walk length (most acute for the log-det barrier near its domain boundary); and \textbf{magnitude dominance} when penalties act on weighted entries so large coefficient products steer gradients. DAGMA's inverse matrix provides a weighted walk-sum closure within its spectral domain, but its published acyclicity objective does not directly supervise individual closure entries---they could, in principle, feed a new pairwise loss, but DAGMA uses them only through the barrier gradient. SMNC instead constructs gate-based, scale-normalized relation scores and uses diagonal and optional off-diagonal entries as explicit supervision targets.

Our main conceptual contributions are threefold. \textbf{First}, \emph{SMNC} exposes explicit matrix-valued structural relation scores: instead of using closure information only implicitly through the derivative of a scalar acyclicity function, it computes soft multi-hop reachability logits on a structural gate \(G\)---one score per ordered pair---for diagonal cycle suppression and optional off-diagonal relation supervision (Section~\ref{sec:wg-factorized}). \textbf{Second}, SMNC normalizes path scores: it propagates scores in log-gate space by repeated dyadic squaring (\(\lceil\log_2 d\rceil\) scales cover paths up to length \(d\)) and divides each scale by its path-length budget before readout, counteracting product-style attenuation on sufficiently strong long cycles; in the low-temperature limit a soft max-plus scan emphasizes strongest paths rather than pooling every walk length into one sum. \textbf{Third}, we \textbf{separate structure from magnitude} so closure supervision acts on support-level gates while data fit, sparsity, and the DAGMA barrier remain on the effective adjacency \(A\): in the linear model, \(A=W_f(M)\odot G(Z)\) with bounded \(W_f\); in the nonlinear extension, an induced gate \(G^{\mathrm{nl}}_{ij}=a_{ij}/(a_{ij}+c)\) yields \(a_{ij}\ge\alpha\Rightarrow G^{\mathrm{nl}}_{ij}\ge\alpha/(\alpha+c)\). The requirement is that strong effective edges imply sufficiently strong structural gates; bounded WG factorization is one sufficient mechanism in the linear setting, not a universal prerequisite for SMNC.

Closure scores are meaningless on a raw weighted adjacency that entangles existence and signed effect size. We therefore keep scalar walk penalties on \(A\) and evaluate SMNC on \(G\) (or \(G^{\mathrm{nl}}\)): data fit, \(\ell_1\) sparsity, and the log-determinant barrier follow DAGMA on \(A\), while \(\mathcal{L}_{\mathrm{cyc}}\) and optional \(\mathcal{L}_{\mathrm{rel}}\) supervise normalized logits built from the gate.

Pairwise domain information usually reports multi-hop precedence (\(i\) reaches \(j\) along some directed path), not one-hop adjacency~\cite{singleintervention_1,temporalorder_1}. We therefore use such cues as targets for closure logits. Used as edge labels, they would encourage the model to place direct edges on transitively related pairs, an error we call transitive overshoot.

We evaluate two versions of the model. The cue-free version uses only the diagonal cycle term and produces all synthetic benchmark results (Table~\ref{tab:main_raw_d100_summary}). The cue version also gives the off-diagonal closure logits pairwise reachability cues as closure targets. A cue \(i\leadsto j\) constrains whether a directed path exists, which edge-level acyclicity penalties cannot express. On the Sachs data (Section~\ref{sec:sachs}), reference-filtered cues improved recovery. The benefit depends on cue accuracy, and noisier cues derived from data alone remain to be tested at scale.

\paragraph{Contributions.}
1) We introduce SMNC, a matrix-valued relation head with explicit per-pair closure logits on a structural gate, based on differentiable soft semiring propagation in log-gate space. 2) We normalize dyadic path scores by scale budgets and establish a cycle-length-independent lower bound on normalized logits for sufficiently strong cycles (Theorem~\ref{thm:length_stable_cycle}); under a unique-active-cycle assumption, the hard final-scale readout admits a polynomially scaled log-gate gradient contribution rather than exponential product-style decay (Corollary~\ref{cor:poly_gradient_decay_bce}; scope in Section~\ref{sec:smnc-properties}). 3) We separate structure from magnitude for closure supervision---bounded \(A=W_f(M)\odot G(Z)\) in the linear model and an induced ratio gate \(G^{\mathrm{nl}}\) in the nonlinear extension---while retaining a DAGMA-style scalar barrier on \(A\). 4) We let off-diagonal logits take optional pairwise reachability cues as closure targets; on Sachs, reference-filtered cues improved recovery in a single-seed run. 5) We evaluate on unstandardized linear and nonlinear synthetic benchmarks and on Sachs, separating cue-free runs from oracle- or reference-filtered side information.

\paragraph{Reachability semantics and cue interface.}
\label{sec:notation}
\label{subsec:closure}
\label{subsec:pairwise-cues}
Write $i\leadsto j$ if a directed path of length $\ge 1$ exists and set $\bar R_{ij}(B)=\mathbf{1}\{i\leadsto j\}$. Then $\mathrm{diag}(\bar R(B))=0$ iff $B$ is acyclic. For DAGs use strict $R(B)=\bar R(B)\circ(\mathbf{1}-I)$ so $i\prec j\Leftrightarrow R_{ij}(B)=1$. Positive reachability labels constrain closure logits, not direct edges (avoid transitive-closure bias). Optional noisy cues are masked binary targets $(O^{(o)},M^{(o)})$ on off-diagonals for Backward-Exclusion (BEx), Ancestry (Anc), and Incomparability (Inc) semantics, closest in spirit to ancestral and side-constraint approaches but implemented here as soft BCE losses on differentiable reachability logits rather than hard feasibility constraints~\cite{chen2016ancestral,li2018sideconstraints,differentialorder,ordereffective_4}.

Even when pairwise reachability agrees, DAGs can differ in the hop scale at which relations first appear, motivating scale-aware multi-scale closure readouts beyond binary ancestry.

\section{Related Work}
\label{sec:related}

\paragraph{Classical causal discovery.}
Classical causal discovery offers several complementary families: constraint-based conditional-independence tests, score-based combinatorial search, hybrid procedures, and order/permutation formulations~\cite{spirtes2000causation,chickering2002optimal,heckerman1995learning,tsamardinos2006maxmin,gasse2014hybrid,raskutti2018sparsest,solus2021greedy,heinze2018causal,kitson2023survey,squires2020permutation,huang2011kdd,huang2013sbn}. These methods establish strong algorithmic and statistical foundations, but they typically rely on discrete search spaces whose exact optimization is difficult at scale. This motivates differentiable surrogates that retain global structure while enabling gradient-based learning.

\paragraph{Continuous DAG optimization.}
Continuous DAG optimization methods replace explicit DAG search with smooth objectives over weighted adjacencies~\cite{notears,nofears,changswap,dagma}. Likelihood-centric variants such as GOLEM pursue alternative graphical-model assumptions~\cite{benchmark_golem}. COSMO and SDCD leave that weighted-walk penalty; the next paragraph states how~\cite{benchmark_cosmo,benchmark_sdcd}.

\paragraph{Walk-based acyclicity constraints.}
NOTEARS-style constraints are built from matrix functions whose powers encode weighted walks and closed walks on a weighted adjacency~\cite{notears,nofears}. DAGMA makes this perspective explicit and replaces trace-exponential or polynomial acyclicity functions by a log-determinant characterization, motivated in part by improved detection of large cycles and better-behaved gradients~\cite{dagma}. The inverse inside the barrier is also a closure: \((sI-B)^{-1}=s^{-1}\sum_{k\ge0}(B/s)^k\), \(B=A\circ A\), sums the weighted walks between every ordered pair whenever the spectral radius of \(B\) is below \(s\), and the gradient on edge \((i,j)\) is \(2A_{ij}[(sI-B)^{-1}]_{ji}\), the weighted sum of the walks from \(j\) back to \(i\) that would close a cycle through that edge. DAGMA's acyclicity objective does not directly supervise individual resolvent entries as prediction targets: they enter only through the barrier gradient, sum walks of every length (Appendix~\ref{app:walk_accounting}), and diverge as the spectral radius approaches \(s\). SMNC instead builds gate-based, scale-normalized relation scores and supervises their diagonal and optional off-diagonal entries explicitly (Section~\ref{sec:wg-factorized}), while still combining closure losses with a scalar DAGMA barrier on \(A\). Scalar polynomial surrogates built from \(\mathrm{tr}(B^\ell)\) recount the same simple cycle at every multiple walk length and impose length dependence through the coefficient schedule \(\{c_\ell\}\) (Appendix~\ref{app:walk_accounting}). SDCD traces training difficulties to instability in acyclicity constraints and motivates a more stable spectral constraint together with a two-stage sparse training procedure~\cite{benchmark_sdcd}; the constraint is a single scalar, and its gradient is supported only on the dominant strongly connected component. COSMO avoids evaluating a cubic acyclicity constraint by parameterizing a smooth acyclic orientation from node priorities and annealing its temperature, while noting that orientation gradients can saturate when priority differences become large~\cite{benchmark_cosmo}.

\paragraph{Nonlinear gradient methods.}
Nonlinear and broader gradient-based formulations extend continuous discovery beyond linear SEM parameterizations. Representative examples include nonlinear NOTEARS, GraN-DAG, DAG-GNN, NoCurl, DCDI, SAM, and RL-based search~\cite{zheng2020sparseNonparametric,lachapelle2020grandag,dag_1,yu2021nocurl,brouillard2020dcdi,kalainathan2022sam,zhu2020rl}. In parallel, identifiability and convergence analyses for nonlinear or non-Gaussian settings clarify when such estimators can recover causal structure and where failure modes arise~\cite{shimizu2006lingam,nonlinearorder_1,nonlinearorder_2,buehlmann2014cam,peters2017elements,ng2022convergence,deng2023global,deng2024markov,ng2024sober,reisach2021beware}.

\paragraph{Side information.}
Side-information and relation-constrained approaches incorporate intervention-derived pairwise relations, temporal order cues, or explicit structural constraints~\cite{singleintervention_1,temporalorder_1,temporalorder_2,temporalorder_3,chen2016ancestral,li2018sideconstraints,differentialorder,ordereffective_4,cao2018expertbn}. Related continuous and time-series frameworks similarly leverage order and pairwise structure~\cite{pamfil2020dynotears,runge2019pcmci,raskutti2018sparsest,solus2021greedy}. Our method also uses pairwise relations as supervision, but applies them to closure logits (multi-hop reachability) rather than to direct-edge labels.

\section{Method}
\label{third:method}

\subsection{Factorized adjacency and closure-side relation heads}
\label{sec:wg-factorized}
\label{sec:reach-layer}

\paragraph{Factorized effective adjacency.}
Separating signed magnitudes \(W_f(M)\) from a structural gate \(G(Z)\) prevents attributing large \(\lvert A_{ij}\rvert\) to vanishing gate mass, which would weaken penalties on graph support and decouple supervision from the closure branch (the \(G\)-side computation feeding the relation heads below, as opposed to the magnitude branch \(W_f\)).
\begin{equation}
\label{eq:A_WG}
A=(W_f(M)\circ G(Z))\circ(\mathbf{1}-I),
\quad
W_f(M)=\beta\tanh(M),
\quad
G(Z)=\sigma(Z/\tau_g)\circ(\mathbf{1}-I),
\end{equation}
Here \(W_f(M)\) stores signed magnitudes, \(G(Z)\) stores soft structural support with off-diagonal entries in \((0,1)\) and zero diagonal, \(M,Z\in\mathbb{R}^{d\times d}\), \(\beta>0\) is a magnitude bound, and \(\tau_g>0\) is a gate temperature; data fit, sparsity, and the DAGMA barrier are evaluated on \(A\), while closure-side cycle and relation losses are applied to \(G\). We abbreviate this factorized backbone as WG (bounded signed-magnitude branch \(W_f\) + structural gate \(G\)).
Since \(|W_{f,ij}|\le\beta\), any effective edge satisfying \(|A_{ij}|\ge\alpha\), for \(0<\alpha\le\beta\), necessarily has \(G_{ij}\ge\alpha/\beta\).
Consequently, every directed cycle in the \(\alpha\)-thresholded effective graph corresponds to a gate cycle whose edge weights are at least \(\alpha/\beta\).
The converse need not hold: a large gate value does not guarantee a large effective edge weight, and when \(W_{f,ij}=0\) one has \(|A_{ij}|=0\) regardless of \(G_{ij}\).
The factorization therefore yields a one-way link from strong effective edges to gates, not a full correspondence between thresholded \(A\) and \(G\).
Positive reachability cues act on closure of \(G\) and need not be realized by corresponding paths in a thresholded effective graph; that mismatch is a limitation of the surrogate, not a flaw in the cycle lower bound (Theorem~\ref{thm:length_stable_cycle}).

Following Section~\ref{sec:notation}, relation heads act on the structural gate \(G\) and produce matrix-valued logits for self-reachability and off-diagonal reachability. The main construction defines the SMNC closure head (a specific relation head, instantiated below) used in our experiments.

\paragraph{Dyadic closure scan.}
Closure-side training needs matrix-valued reachability summaries on the structural gate \(G\): diagonal logits encode self-reachability and directed cycles (for suppression), while optional off-diagonal logits support masked reachability-style BCE, addressing relation targets such as ancestry that scalar acyclicity penalties on the effective adjacency \(A\) in Eq.~\eqref{eq:A_WG} cannot capture alone.
Our dyadic states \(U^{(k)}_{ij}\) supply these logits in softened form: computed from \(\log G\), each aggregates multi-hop connectivity by scoring the \emph{best} directed explanation from \(i\) to \(j\) among routes of length at most \(2^k\), complementing the edge-level parameterization on \(A\) above.
Naively materializing all walk lengths for all pairs does not scale; we maintain a compact sufficient-statistics stack \(\{U^{(k)}\}\) indexed by dyadic path-length budgets, with repeated doubling instead of explicit path enumeration.
Classical directed reachability is packaged by repeated matrix composition in the Boolean semiring \((\lor,\land)\)~\cite{warshall1962boolean}, and graph path problems more generally admit semiring/dioid formulations~\cite{gondran2008graphs}.

Training pairs logits \(S_{ij}\) with bounded scores \(P_{ij}=\sigma(S_{ij})\) (concretely \(P^{\mathrm{un}}_{ij}\) and \(P^{(\gamma)}_{ij}(\Omega)\) in the SMNC readout below; BCE supervision in Section~\ref{sec:unified-bce}).
Along a directed route \(\pi:i\leadsto j\), whole-route openness is \(\prod_{e\in\pi} G_e\); with log-gates \(E_{ij}=\log G_{ij}\), \(\sum_{e\in\pi} E_e\) turns products into sums, so best-route summaries are \emph{max--sum} in \(E\)-space (equivalently max--\emph{product} on gates).
We approximate \(\max_{\pi:\,i\leadsto j,\,|\pi|\le L}\sum_{e\in\pi} E_e\) for all pairs and budgets without enumerating paths using the dyadic scan in Eq.~\eqref{eq:smnc-hard-recursion}; Theorem~\ref{thm:dyadic_semantics} records the exact hard-\((\vee,\otimes)\) semantics.
Readouts map the resulting stack \(\{U^{(k)}\}\) to the logits \(S_{ij}\) used in training (e.g.\ \(\Lambda_{ij}=U^{(K)}_{ij}\) in Eq.~\eqref{eq:unnorm-dyadic-logit} and the SMNC readouts below).

Formally, the log-gates are \(E_{ij}=\log G_{ij}\) for \(i\neq j\) and \(E_{ii}=-\infty\), and we set \(K=\lceil \log_2 d\rceil\). Thus explicit self-edges are forbidden, while self-reachability may still arise through multi-step directed cycles.
The hard \((\max,+)\) dyadic scan is repeated squaring in the max-plus semiring, a standard path-algebraic construction~\cite{gondran2008graphs}:
\begin{equation}
\label{eq:smnc-hard-recursion}
U^{(0)}=E,\qquad U^{(k+1)}=U^{(k)}\vee\bigl(U^{(k)}\otimes U^{(k)}\bigr),\qquad k=0,\ldots,K-1,
\end{equation}
where \((a\vee b)=\max\{a,b\}\) and \((X\otimes Y)_{ij}=\max_m(X_{im}+Y_{mj})\).
For training, hard maxima are replaced by log-sum-exp surrogates at closure temperature \(\tau_c>0\), preserving the dyadic doubling pattern of Eq.~\eqref{eq:smnc-hard-recursion} while routing gradients through competing path explanations.
We use soft max-plus (Eq.~\eqref{eq:smnc-soft-recursion}) for autodiff; concretely, \(a\vee_{\tau_c}b=\tau_c\log\!\bigl(\exp(a/\tau_c)+\exp(b/\tau_c)\bigr)\) and \((X\otimes_{\tau_c}Y)_{ij}=\tau_c\log\sum_m\exp\!\bigl((X_{im}+Y_{mj})/\tau_c\bigr)\).
Training implements each composition as a shifted exponential-space GEMM in \texttt{float32} (linear) or a clamped approximate GEMM (nonlinear); these match the exact scan only where finite precision preserves the products (Appendices~\ref{app:dyadic_scan_unpack} and~\ref{app:nonlinear_induced_g}).
The semiring semantics and gradient envelopes below concern exact arithmetic unless stated otherwise.
The same primitives extend to general (non-dyadic) length budgets in two ways. One composes additional segment max-plus products with step sizes other than pure doubling, which needs more rounds than the \(O(\log d)\) depth of repeated squaring. The other pads the cap upward to the next dyadic tier.
\begin{equation}
\label{eq:smnc-soft-recursion}
U^{(0)}=E,\qquad
U^{(k+1)}=U^{(k)}\vee_{\tau_c}\bigl(U^{(k)}\otimes_{\tau_c} U^{(k)}\bigr),
\qquad k=0,\dots,K-1.
\end{equation}

\begin{theorem}[Exact semantics of the hard dyadic scan]
\label{thm:dyadic_semantics}
For every \(k\ge 0\),
\(
U^{(k)}_{ij}
=
\max\{\sum_{e\in\pi} E_e:\ \pi\text{ a directed walk from }i\text{ to }j,\ 1\le |\pi|\le 2^k\}
\)
(\(\max\emptyset=-\infty\)). Because \(E\le 0\), whenever the walk set is nonempty the maximum is attained by a simple path if \(i\ne j\) and by a simple cycle through \(i\) if \(i=j\).
\end{theorem}

The proof is given in Appendix~\ref{app:proof-dyadic-semantics}.

In the hard scan of Eq.~\eqref{eq:smnc-hard-recursion}, \(U^{(0)}=E\) stores one-hop \(\log G\) scores. The product \(U^{(0)}\otimes U^{(0)}\) maximizes over intermediate \(m\) and thus collects the best concatenation of two length-\(\le 1\) segments, i.e.\ the best length-\(\le 2\) path in log-space; \(U^{(1)}=U^{(0)}\vee(U^{(0)}\otimes U^{(0)})\) keeps whichever one-hop or two-hop route is stronger. Inductively, \(U^{(k)}\) carries the best score among paths of length at most \(2^k\). Only \(K=\lceil\log_2 d\rceil\) repeated-squaring rounds are needed to reach budgets \(2^K\ge d-1\) along simple routes, hence \(O(\log d)\) depth and no path enumeration.

Each update merges a best length-\(\le 2^k\) explanation from \(i\) to an intermediate \(m\) with a best length-\(\le 2^k\) explanation from \(m\) to \(j\), then maximizes over splits \(m\). Standard matrix multiplication computes \(\sum_m (A_{im} B_{mj})\). By operating in the \((\max,+)\) (max-plus) semiring, we simply replace addition with maximization and multiplication with addition, computing \(\max_m (A_{im} + B_{mj})\). In the hard (\(\tau_c\to0\)) limit this repeated squaring isolates the single strongest route; the smooth version used for training is discussed below.

All relation heads share the states \(U^{(k)}\) of the soft scan in Eq.~\eqref{eq:smnc-soft-recursion} and differ only in their readout, the map from \(\{U^{(k)}\}_{k=0}^K\) to relation logits.

\paragraph{SMNC: scale-normalized multi-scale closure head.}
The simplest dyadic readout maps the final raw dyadic closure state to relation logits,
\begin{equation}
\label{eq:unnorm-dyadic-logit}
\Lambda_{ij}:=U^{(K)}_{ij},
\qquad
P^{\mathrm{un}}_{ij}:=\sigma(\Lambda_{ij}).
\end{equation}
This baseline preserves the structure/magnitude separation and closure-side supervision principle, but it does not scale-normalize path scores. On a unique length-\(L\) path with constant gate value \(p\in(0,1)\), the dominant raw logit scales as \(L\log p\). 

The name \emph{Scale-Normalized Multi-Scale Closure} (SMNC) refers to (i)~\emph{multi-scale} states \(U^{(k)}\) read at several dyadic scales \(k\), and (ii)~\emph{scale normalization} at each scale: \(\widetilde U^{(k)}=U^{(k)}/2^{\gamma k}\) divides accumulated log-path scores by the length budget \(2^k\) (up to \(\gamma>0\)), counteracting shrinkage of product-form scores along long routes and the associated vanishing BCE gradients (Remark~\ref{rem:old_vs_new_decay}).
\begin{equation}
\label{eq:smnc-normalized-scale}
\widetilde U^{(k)}=\frac{U^{(k)}}{2^{\gamma k}},
\qquad \gamma>0.
\end{equation}
For a selected scale set \(\Omega\subseteq\{0,1,\dots,K\}\), the smooth readout used in practice is
\begin{equation}
\label{eq:smnc-readout-soft}
S_{ij}^{(\gamma)}(\Omega)=
\begin{cases}
\widetilde U_{ij}^{(k)}, & \Omega=\{k\},\\[3pt]
\tau_s\log\sum_{k\in\Omega}\exp\!\bigl(\widetilde U_{ij}^{(k)}/\tau_s\bigr), & |\Omega|>1,
\end{cases}
\qquad \tau_s>0.
\end{equation}
The bounded relation score is
\(
P^{(\gamma)}_{ij}(\Omega)=\sigma\!\left(S^{(\gamma)}_{ij}(\Omega)\right).
\)
We do not interpret \(P^{(\gamma)}_{ij}(\Omega)\) as a calibrated reachability probability. The BCE-with-logits losses below use \(S^{(\gamma)}_{ij}(\Omega)\) as a margin-style surrogate logit for cycle suppression and optional relation supervision.

We use the scale families (fixed subsets \(\Omega\) of dyadic scales \(k\) read out jointly via Eq.~\eqref{eq:smnc-readout-soft})
\(
\Omega_{\mathrm{dir}}=\{0\},\
\Omega_{\le 2}=\{0,1\},\
\Omega_{\le 4}=\{0,1,2\},\
\Omega_{\mathrm{final}}=\{K\},\
\Omega_{\mathrm{all}}=\{0,1,\dots,K\}.
\)
The linear cycle readout \(\Omega_{\mathrm{final}}\) divides by the graph-wide dyadic budget \(2^K=\Theta(d)\), not by an active cycle's length \(L\), so it is not cycle-adaptive path-length normalization (Corollary~\ref{cor:poly_gradient_decay_bce}'s \(\Theta(1/L)\) factor arises only at the minimal covering scale \(k_L\); hard all-scale maximization selects the coarsest scale \(K\) by Lemma~\ref{lem:coarsest-scale}).
These sets expose cumulative scale-normalized relation profiles from the same structural representation. The terms direct, final-scale, and all-scale are shorthand for \(\Omega_{\mathrm{dir}}\), \(\Omega_{\mathrm{final}}\), and \(\Omega_{\mathrm{all}}\). Because \(U^{(k)}\) stores paths of length at most \(2^k\), a shorter path may also contribute to larger-scale readouts, so no family gives exact hop-bin probabilities. Exact hop-bin supervision would require residual or bin-specific readouts, which we leave to future work.

The smooth scan is itself a tempered walk sum, \(\exp(U^{(k)}/\tau_c)=\sum_{\ell\le 2^k}c_{k,\ell}(G^{\circ 1/\tau_c})^{\ell}\) with fixed integers \(c_{k,\ell}\ge1\), and \(0\le U^{(k)}_{\mathrm{soft}}-U^{(k)}_{\mathrm{hard}}\le(2^k-1)\tau_c\log(2d)\) (Lemma~\ref{lem:soft-excess}). The soft scan therefore still aggregates walks, but with weights: each extra traversal of a cycle with gate product \(Q\) is discounted by \(Q^{1/\tau_c}\) and stays in the sum. For short cycles with gates near 1, multiplicity can return. For a 2-cycle with gates \(0.9\) at \(\tau_c=0.1\) and \(d=128\), it multiplies the gradient about 12-fold (Appendix~\ref{app:proof-coarsest-scale}).
\phantomsection\label{sec:unified-bce}
We use BCE-with-logits for diagonal cycle suppression and optional off-diagonal relation cues.
In the linear SMNC model used in our experiments, the cycle logit is the final-scale normalized readout \(S^{\mathrm{cyc}}=S^{(\gamma)}(\Omega_{\mathrm{final}})\), \(\Omega_{\mathrm{final}}=\{K\}\); the nonlinear variant uses \(\Omega_{\mathrm{all}}\) (Section~\ref{sec:smnc-properties}).
The cycle target is zero self-reachability on the diagonal.
Optional off-diagonal cue families, when used, apply masked BCE to \(S^{(\gamma)}_{ij}(\Omega_o)\) with family-specific masks, targets, and weights.
We use BCE as a margin-style relation surrogate rather than a calibrated probability model: hard \((\max,+)\) closure logits are nonpositive, so even an ideally open positive relation incurs loss at least \(\log 2\).
The length-attenuation analysis in Section~\ref{sec:smnc-properties} concerns zero-target cycle suppression, where \(\partial\operatorname{softplus}(s)/\partial s=\sigma(s)\); positive ancestry targets use \(\operatorname{softplus}(-s)\) with \(\partial\ell/\partial s=\sigma(s)-1\), which does not share the same vanishing behavior as \(s\to-\infty\).

\subsection{Length behavior of the scale-normalized head}
\label{sec:smnc-properties}

The results in this section are local and concern a fixed gate matrix and hard \((\max,+)\) readouts unless noted.
Theorem~\ref{thm:length_stable_cycle} is a \emph{score} lower bound: it links thresholded cycles in \(A\) to normalized diagonal logits on \(G\), but does not assert convergence of optimization or gradients along a training trajectory.
Corollary~\ref{cor:poly_gradient_decay_bce} bounds an active-cycle gradient contribution in \(\log g\) under a unique-active-cycle assumption; it is not an unconditional guarantee for the finite-\(\tau_c\) soft scan or for learned parameters \(Z\).
We analyze the hard readout \(S^{(\gamma),\max}_{ij}(\Omega)=\max_{k\in\Omega}U^{(k)}_{ij}/2^{\gamma k}\), with \(U^{(k)}\) from Eq.~\eqref{eq:smnc-hard-recursion}. In exact arithmetic the smooth recursion and scale aggregation dominate their hard counterparts, so hard lower bounds carry over.
The archived linear trainer evaluates the soft scan with row/column-shifted \texttt{float32} GEMMs; that implementation can zero out logits whose best compositions lie far below the shifts, invalidating the cycle bounds and gradients on affected instances (Appendix~\ref{app:dyadic_scan_unpack}).
The nonlinear trainer uses a further clamped GEMM (Section~\ref{sec:reach-regularized}); it is an approximate implementation to which the hard-scan and exact soft-scan theorems do not apply directly.

\begin{theorem}[Length-stable cycle lower bound at \(\gamma=1\)]
\label{thm:length_stable_cycle}
Assume the bounded backbone in Eq.~\eqref{eq:A_WG}. Suppose the thresholded
effective graph contains a directed cycle \(C\) of length \(L\), and every edge
\(e\in C\) satisfies \(|A_e|\ge \alpha\), with \(0<\alpha\le \beta\). Let
\(k_L:=\lceil\log_2 L\rceil\) and \(K=\lceil\log_2 d\rceil\). Then every node
\(v\) on the cycle satisfies
\(
\frac{U_{vv}^{(k_L)}}{2^{k_L}}
\ge
\log(\alpha/\beta),
\qquad
\frac{U_{vv}^{(K)}}{2^K}
\ge
\log(\alpha/\beta).
\)
Consequently, for both readouts the normalized self-reachability logit of a
thresholded cycle is bounded below independently of \(L\), whereas the raw logit
\(U^{(K)}_{vv}\) is only bounded below by \(L\log(\alpha/\beta)\).
\end{theorem}
The proof is given in Appendix~\ref{app:proof-length-stable-cycle}.
\begin{corollary}[Scale-specific polynomial gradient factor]
\label{cor:poly_gradient_decay_bce}
Assume the hard normalized readout with \(\gamma=1\). Let \(C\) be a cycle
candidate of length \(L\), and consider a selected dyadic scale \(a\) such that the cycle candidate is feasible at that scale. Define
\(
S_{vv,a}:=\frac{U^{(a)}_{vv}}{2^a}.
\)
If the cycle candidate is the unique maximizer in \(U^{(a)}_{vv}\), then for
every edge \(e\) on this active candidate,
\(
\frac{\partial S_{vv,a}}{\partial \log g_e}
=
2^{-a}.
\)
Consequently, for the zero-target BCE cycle loss \(\ell^{\rm cyc}_{v,a}=\operatorname{softplus}(S_{vv,a})\),
\(
\frac{\partial \ell^{\rm cyc}_{v,a}}
{\partial \log g_e}
=
2^{-a}\sigma(S_{vv,a}).
\)
At the minimal covering scale \(a=k_L=\lceil\log_2 L\rceil\), this gives a \(\Theta(1/L)\) scale factor. At the final scale \(a=K=\lceil\log_2 d\rceil\), which is the readout used by the linear cycle loss, the factor is
\(\Theta(1/d)\).
\end{corollary}
These bounds concern local derivatives with respect to log-gates; gradients in the trained gate logits additionally include the sigmoid Jacobian \((1-g_e)/\tau_g\) and need not obey the same lower bound (Appendix~\ref{app:proof-coarsest-scale}).
Proofs and a discussion of active scales in all-scale readouts are given in
Appendix~\ref{app:proof-coarsest-scale}.

Lemma~\ref{lem:coarsest-scale} (stated and proved in Appendix~\ref{app:proof-coarsest-scale}) gives
coarsest-scale dominance of the hard readout: for any \(\gamma>0\), \(U^{(a)}_{ij}/2^{\gamma a}\) is
non-decreasing in \(a\), so the hard \(\Omega_{\mathrm{all}}\) and \(\Omega_{\mathrm{final}}\) readouts coincide.
Off-diagonal soft normalized scores need not be monotone in \(a\) (Lemma~\ref{lem:soft-excess}), whereas exact soft diagonal states satisfy the same non-decreasing property at \(\gamma=1\) (Appendix~\ref{app:proof-coarsest-scale}).
Smooth scale aggregation (\(\tau_s>0\)) spreads gradient over finer scales.

\paragraph{Objective used in our experiments.}
Linear benchmarks use \(\mathcal L_{\mathrm{cyc}}=\frac1d\sum_v\operatorname{softplus}(U^{(K)}_{vv}/2^K)\)
(\(\Omega_{\mathrm{final}}\), \(\gamma=1\), \(\tau_c=0.1\)); nonlinear ones use the soft
\(S^{(1)}(\Omega_{\mathrm{all}})\), \(\tau_c=\tau_s=0.1\) (with a clamped GEMM, Appendix~\ref{app:nonlinear_induced_g}). For a thresholded length-\(L\) cycle \(C\) that uniquely
maximizes \(U^{(K)}_{vv}\), \(v\in C\), Corollary~\ref{cor:poly_gradient_decay_bce} and
Theorem~\ref{thm:length_stable_cycle} give, in the hard limit, a contribution of the \(C\)-terms to
\(\partial\mathcal L_{\mathrm{cyc}}/\partial\log g_e\) (\(e\in C\)) in
\([L\sigma(\log(\alpha/\beta))/(2d^2),\,L/(2d^2)]\), a lower envelope proportional to \(L/d^2\) at fixed \(d\) and active-gate floor; the realized per-edge contribution need not grow monotonically with \(L\) (on a uniform-gate cycle, the hard final-scale factor \(L/(dD)\,\sigma(L\log p/D)\), \(D=2^K\), can decrease when \(p<1\)).
The unnormalized head gives at most
\((L/d)\,\sigma(L\log p_{\max})\), \(p_{\max}:=\max_{e\in C}g_e\) (Appendix~\ref{app:proof-coarsest-scale}).
The \(\Theta(1/L)\) rate holds at the minimal covering scale \(k_L\), while the linear objective reads out the final scale \(K\) with factor \(\Theta(1/d)\).
The bound does not cover the finite-\(\tau_c\) scan exactly, \texttt{float32} shifted-GEMM training, or the nonlinear clamped head.

\subsection{Cycle-regularized continuous DAG learning}
\label{sec:reach-regularized}

For linear Gaussian SEMs,
\[
\mathrm{score}(A)=\frac{1}{2n}\lVert X-XA\rVert_F^2.
\]
\begin{equation}
\label{eq:dagma-barrier}
h_s(A)=d\log s-\log\det\!\bigl(sI-A\circ A\bigr).
\end{equation}
\begin{equation}
\label{eq:wg-dagma-obj}
\min_{M,Z}\;
\mu\bigl(\mathrm{score}(A)+\lambda_1\lVert A\rVert_1\bigr)
+h_s(A)
+\mu_{\mathrm{cycle}}\mathcal L_{\mathrm{cyc}}
+\mu_{\mathrm{rel}}\mathcal L_{\mathrm{rel}}.
\end{equation}
Setting \(\mu_{\mathrm{rel}}=0\) yields the cue-free closure-regularized model; \(\mu_{\mathrm{cycle}}=\mu_{\mathrm{rel}}=0\) is the pure bounded-WG backbone. In the cue-free model used for the main benchmarks, \(\mu_{\mathrm{rel}}=0\) and
the diagonal cycle term uses the SMNC final-scale readout \(\Omega_{\mathrm{final}}=\{K\}\) in the linear solver (the nonlinear variant below uses \(\Omega_{\mathrm{all}}\) with \(\tau_s=0.1\)).

\paragraph{Nonlinear induced-structure variant.}
A decoder-induced nonnegative \(A_{\mathrm{mag}}\) is mapped to probabilities \(p=\phi(A_{\mathrm{mag}})\circ(\mathbf{1}-I)\) with \(p_{ij}\in[0,1]\); the same closure readouts and \(\mathcal{L}_{\mathrm{cyc}},\mathcal{L}_{\mathrm{rel}}\) are applied to \(p\).
This variant uses a \emph{clamped approximate} soft scan: exponents are clipped before multiplication and products are floored, which can inject spurious path evidence and zero gradients on genuinely weak compositions (Appendix~\ref{app:nonlinear_induced_g}); nonlinear benchmark numbers should be read as empirical outcomes of this head, not as validation of the exact theorems.


\section{Experiments}

\subsection{Toy cyclic examples}
\label{sec:mechan}

\begin{figure}[t]
  \centering
  \includegraphics[width=\linewidth]{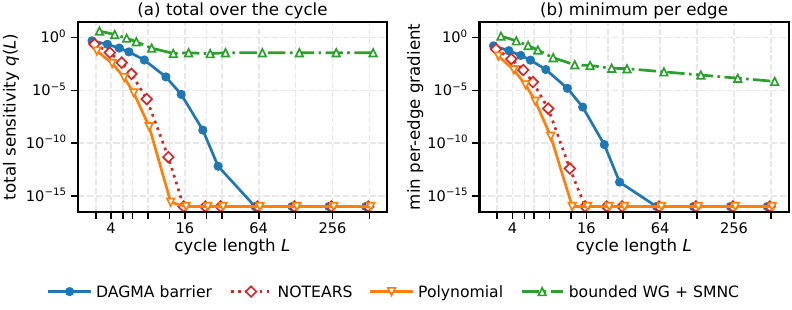}
  \caption{\textbf{Gradient signal on symmetric long cycles.} Matched directed \(L\)-cycles \(X_L(\rho)=\rho B_L\) at \(\rho=0.6\), without optimization, for the walk-sum penalties, namely the DAGMA log-det barrier, the NOTEARS trace-exponential and the DAG-GNN polynomial \(\operatorname{tr}[(I+A\circ A/d)^d]-d\) (all on \(A=X_L(\rho)\)), and for the bounded-WG induced \(A\)-space signal of the DAGMA barrier plus the SMNC final-scale cycle loss (``bounded WG+SMNC'', on \(G=X_L(\rho)\), \(\mu_{\mathrm{cycle}}=1\), \(\tau_g=1\); the barrier dominates for \(L\le8\)). Each cycle is the whole graph (\(d=L\)). (a) Total active-edge sensitivity \(q(L)=\bigl|\sum_{e\in C_L}\nabla_e\bigr|\). (b) Minimum per-edge gradient \(\min_{e\in C_L}|\nabla_e|\), which equals \(q(L)/L\) because every edge of a symmetric cycle receives the same gradient. Values below \(10^{-16}\) are drawn at that floor (Appendix~\ref{app:long_cycle_decay}).}
  \label{fig:c1a_total_main}
\end{figure}

Before the large-scale benchmarks, we run \emph{local probes}: fixed small graphs, no data likelihood, and no search over hyperparameters.
The goal is to separate two questions that scalar walk penalties blur together---whether a penalty still \emph{senses} a long cycle in its gradient, and whether gradient steps can actually \emph{open} that cycle during optimization.

\paragraph{Symmetric long cycles.}
We place a single directed \(L\)-cycle with equal edge strength \(\rho=0.6\) and measure how much decycling signal each objective sends to the cycle edges (Figure~\ref{fig:c1a_total_main}; probe setup in Appendix~\ref{app:long_cycle_decay}).
Walk-sum penalties (DAGMA, NOTEARS, DAG-GNN polynomial) lose total sensitivity \(q(L)\) rapidly as \(L\) grows (log--log slopes \(-76.28\), \(-22.73\), \(-88.03\)).
Bounded WG+SMNC decays far more slowly overall (slope \(-0.78\)), because it still contains the DAGMA barrier; the SMNC cycle term by itself is nearly flat in \(L\) on this isolated-cycle setup.
Spectral penalties such as SDCD are not plotted; the SDCD comparison is in Appendix~\ref{app:nonwalk}.

\paragraph{Controlled single- and two-cycle optimization.}
Figure~\ref{fig:c1a_total_main} evaluates gradients at a fixed graph.
The next probe runs short full-gradient optimizations on toy cyclic supports: only cycle edges are free, the adjacency is \(A=G\), and the objective is a fidelity term plus DAGMA's log-det barrier and either a raw or an SMNC cycle loss (no data term).
We mark a cycle as broken when any gate on it falls below \(0.1\); setup and hyperparameters are in Appendix~\ref{app:controlled-long-cycle}.

We first match the \emph{initial} gradient on the closing edge at \(L=4\), then freeze those cycle weights and vary \(L\).
Under that calibration, SMNC maintains stronger pressure on long cycles than the unnormalized head as \(L\) grows.
At one fixed pair of weights for both heads, the unnormalized head clears short single cycles (\(L\le 8\)) within 60 updates but does not break any \(L\ge 16\) cycle within 500 updates.
SMNC breaks \(L\in\{16,24,32\}\), and does so in fewer updates as \(L\) increases (Figure~\ref{fig:time-to-break}), while retaining most path-gate mass on the surviving edges.

We then embed a short cycle (\(L_s=8\)) and a long one on the same support with one shared cycle weight.
SMNC breaks the long cycle first in every run, but the short cycle lags: joint removal of both cycles often waits until the weaker short cycle finally opens.
An unnormalized closure can finish both cycles faster only when the long cycle is relatively short (\(L_\ell=24\)) and never clears the longest cycles within the tested budget, even with up to 5000 updates.
Appendix~\ref{app:controlled-long-cycle} also reports sufficient cycle weights on isolated cycles and an \(s\)-continuation control that weakens the log-det barrier stage by stage.

\subsection{Synthetic Gaussian benchmarks}
\label{sec:synth}

\paragraph{Models and data.}
We report two cue-free variants: the explicit bounded-WG linear solver (Section~\ref{sec:wg-factorized}) and the nonlinear induced-structure extension (Section~\ref{sec:reach-regularized}).
Data are drawn from Erd\H{o}s--R\'enyi DAGs with \(d\in\{10,15,20\}\cup\{30,40,\ldots,100\}\) and \(s_0=\kappa d\) for \(\kappa\in\{2,4,5,6,7,8\}\) when feasible.
For each \((d,s_0)\) pair we sample \(N_{\mathrm{rep}}=5\) independent datasets of \(n=10{,}000\) observational points.
Linear SEMs use edge weights from \(\mathrm{Unif}([-1.5,-0.5]\cup[0.5,1.5])\) and homoscedastic Gaussian noise with \(\sigma=0.6\).
Nonlinear SEMs use width-100 one-hidden-layer MLP mechanisms in topological order.
Following the usual NOTEARS-style benchmark protocol, inputs are not standardized (Limitations).

\paragraph{Cue-free training setup.}
Main benchmark rows use diagonal closure-side cycle regularization only: no off-diagonal relation BCE and no hard reverse-edge mask.
Linear runs read out \(\Omega_{\mathrm{final}}=\{K\}\) via shifted \texttt{float32} GEMM; nonlinear runs use the soft all-scale readout \(\Omega_{\mathrm{all}}\) at \(\tau_s=0.1\) with the clamped GEMM (Sections~\ref{sec:unified-bce} and~\ref{sec:smnc-properties}).
Reported linear gains are therefore under the documented numerical closure, not the idealized exact scan assumed in Section~\ref{sec:smnc-properties}.

\paragraph{Baselines and reporting.}
We compare against DAGMA, NOTEARS, GOLEM, COSMO, and SDCD.
The main text emphasizes DAGMA as the closest single-adjacency continuous baseline; NOTEARS (DAGrad's default Adam run) is consistently worse than both DAGMA and SMNC in all 61 \((d,s_0)\) settings, under RAW and TOPK and on both SEM types.
Full \(d=100\) tables and across-\(d\) curves appear in Appendix~\ref{app:overview}; baseline budgets and hyperparameters were not matched to ours and are disclosed in Appendix~\ref{app:baseline-protocol}.

\paragraph{Metrics and optimizer schedule.}
\label{sec:metrics}
We score structure recovery on the learned effective adjacency \(A\), not on closure logits, using pair-space SHD and directed edgewise SHD (dSHD); reversing an edge counts as two edits.
We report both \textbf{RAW} thresholded outputs and \textbf{TOPK} outputs that keep up to \(s_0\) largest-magnitude edges (a ranking diagnostic with known sparsity).
Table~\ref{tab:main_raw_d100_summary} gives RAW summaries at \(d=100\); TOPK and across-\(d\) curves are in Appendix~\ref{app:overview}.
For the linear bounded-WG solver we use a DAGMA-style outer loop with \(N_{\mathrm{outer}}=5\), \(\mu_{\mathrm{init}}=10\), stagewise \(s\in\{1.0,0.9,0.8,0.7,0.6\}\), \(\lambda_1=10^{-2}\), \(\tau_g=1\), magnitude threshold \(0.1\), and 10{,}000 or 20{,}000 inner steps per stage (depending on \(d\) and ER).
The backbone runs in \texttt{float64} and the closure head in \texttt{float32} (shifted GEMM; Appendix~\ref{app:dyadic_scan_unpack}). Archived reruns reproduce every reported cell to within one standard deviation (Appendix~\ref{app:baseline-protocol}).

\paragraph{Cue-free \(d=100\) results.}
\label{sec:main-benchmark-summaries}

\begin{table*}[t]
    \centering
    \scriptsize
    \setlength{\tabcolsep}{2.8pt}
    \caption{Cue-free RAW results at \(d=100\). Each cell reports mean $\pm$ sample standard deviation over 5 datasets.}
    \label{tab:main_raw_d100_summary}
    \resizebox{0.90\textwidth}{!}{%
    \begin{tabular}{lcccccccc}
        \toprule
        & \multicolumn{4}{c}{Gaussian linear SEMs} 
        & \multicolumn{4}{c}{Gaussian nonlinear SEMs} \\
        \cmidrule(lr){2-5}\cmidrule(lr){6-9}
        & \multicolumn{2}{c}{bounded WG + SMNC}
        & \multicolumn{2}{c}{DAGMA}
        & \multicolumn{2}{c}{induced + SMNC}
        & \multicolumn{2}{c}{DAGMA} \\
        \cmidrule(lr){2-3}\cmidrule(lr){4-5}
        \cmidrule(lr){6-7}\cmidrule(lr){8-9}
        ER 
        & dSHD $\downarrow$ & F1-dir $\uparrow$
        & dSHD $\downarrow$ & F1-dir $\uparrow$
        & dSHD $\downarrow$ & F1-dir $\uparrow$
        & dSHD $\downarrow$ & F1-dir $\uparrow$ \\
        \midrule
        2
        & 2.20 $\pm$ 3.19 & \textbf{0.99 $\pm$ 0.01}
        & \textbf{2.00 $\pm$ 3.08} & \textbf{0.99 $\pm$ 0.01}
        & 69.80 $\pm$ 23.86 & \textbf{0.83 $\pm$ 0.05}
        & \textbf{68.80 $\pm$ 23.53} & 0.82 $\pm$ 0.06 \\
        4
        & \textbf{22.80 $\pm$ 17.05} & \textbf{0.97 $\pm$ 0.02}
        & 48.00 $\pm$ 41.36 & 0.94 $\pm$ 0.05
        & \textbf{85.40 $\pm$ 11.28} & \textbf{0.89 $\pm$ 0.01}
        & 116.60 $\pm$ 8.08 & 0.84 $\pm$ 0.01 \\
        6
        & \textbf{20.80 $\pm$ 15.01} & \textbf{0.98 $\pm$ 0.01}
        & 88.20 $\pm$ 47.33 & 0.93 $\pm$ 0.04
        & \textbf{127.00 $\pm$ 23.40} & \textbf{0.89 $\pm$ 0.02}
        & 204.20 $\pm$ 6.02 & 0.80 $\pm$ 0.01 \\
        7
        & \textbf{58.80 $\pm$ 22.17} & \textbf{0.96 $\pm$ 0.01}
        & 102.20 $\pm$ 47.01 & 0.93 $\pm$ 0.03
        & \textbf{173.60 $\pm$ 28.08} & \textbf{0.87 $\pm$ 0.02}
        & 290.60 $\pm$ 12.66 & 0.75 $\pm$ 0.02 \\
        8
        & \textbf{248.40 $\pm$ 103.58} & \textbf{0.84 $\pm$ 0.06}
        & 376.80 $\pm$ 206.54 & 0.78 $\pm$ 0.11
        & \textbf{247.00 $\pm$ 36.86} & \textbf{0.82 $\pm$ 0.03}
        & 390.40 $\pm$ 27.61 & 0.69 $\pm$ 0.03 \\
        \bottomrule
    \end{tabular}}
\end{table*}

\paragraph{Linear SEMs.}
\label{sec:linear-results}
On linear Gaussian SEMs, Table~\ref{tab:main_raw_d100_summary} shows bounded WG + SMNC competitive with or better than DAGMA in denser regimes, reducing dSHD from 48.0 to 22.8 at ER=4, from 88.2 to 20.8 at ER=6, from 102.2 to 58.8 at ER=7, and from 376.8 to 248.4 at ER=8, while maintaining high directed F1.
At ER=2, all methods are close under RAW evaluation.

\paragraph{Nonlinear SEMs.}
\label{sec:nonlinear-results}
On nonlinear Gaussian SEMs, Table~\ref{tab:main_raw_d100_summary} shows the induced-structure variant with the same pattern: it improves over DAGMA in denser regimes (e.g.\ dSHD 116.6 vs.\ 85.4 at ER=4, 390.4 vs.\ 247.0 at ER=8) and matches DAGMA closely at ER=2.
Our nonlinear solver also differs from the DAGMA baseline outside the cycle term (wider MLP, extra outer stage, different \(\lambda_1\) and \(\mu_{\mathrm{init}}\)), so this comparison does not isolate SMNC alone. Table~\ref{tab:soft_cue_nonlinear} reports a nonlinear ablation of hard exclusion and soft reachability supervision using oracle ancestry cues (Appendix~\ref{app:hard-mask}).

\subsection{Real-data validation: Sachs}
\label{sec:sachs}
We evaluate on the Sachs protein-signaling data~\cite{sachs} against the standard 17-edge reference DAG distributed with \texttt{bnlearn} (URL in Appendix~\ref{app:sachs_order_cues}), a literature benchmark and not an absolute biological ground truth. We report single-seed (seed\(=0\)) RAW recovery at output threshold \(0.3\) on 11{,}672 pooled, column-wise z-scored samples from all 14 conditions. The first row of Table~\ref{tab:sachs_17edge_raw} is the internal nonlinear backbone with the SMNC cycle term disabled; all other rows keep cycle regularization. The cue-assisted rows share 12 ancestry relations extracted from interventional shifts and filtered with the reference DAG (Appendix~\ref{app:sachs_order_cues}). They compare hard reverse-edge exclusion alone, positive-only soft closure supervision without a hard mask, and hard exclusion with two-sided soft supervision (positive forward and zero reverse targets).


\begin{table}[t]
\centering
\small
\setlength{\tabcolsep}{3pt}
\caption{Sachs RAW recovery against the 17-edge reference DAG, using seed 0 and output threshold $t=0.3$. The first row is the internal nonlinear backbone with the cycle term disabled, not a separately executed official DAGMA baseline. The target column lists the numbers of active positive and negative soft closure targets; ``--'' denotes no active soft loss. }
\label{tab:sachs_17edge_raw}
\begin{tabular*}{\linewidth}{@{\extracolsep{\fill}}lccrr@{}}
\toprule
Configuration & \shortstack{Hard\\mask} & \shortstack{Soft targets\\pos./neg.} & dSHD $\downarrow$ & F1-dir $\uparrow$ \\
\midrule
Nonlinear backbone (cycle off) & No & -- & 24 & 0.250 \\
Induced + SMNC, no cues & No & -- & 14 & 0.533 \\
SMNC + hard exclusion & Yes & -- & 14 & 0.500 \\
SMNC + positive-only soft cues & No & $12/0$ & 10 & 0.722 \\
SMNC + hard + two-sided soft cues & Yes & $12/12$ & \textbf{6} & \textbf{0.833} \\
\bottomrule
\end{tabular*}
\end{table}

Enabling the cycle term lowers dSHD from 24 to 14. Hard exclusion alone leaves dSHD at 14 and lowers F1-dir from 0.533 to 0.500, positive-only soft supervision reaches dSHD 10 and F1-dir 0.722 without hard exclusions, and hard plus two-sided soft supervision reaches dSHD 6 and F1-dir 0.833. These single-seed results suggest that soft reachability supervision can complement relations the cue-free model does not reliably recover. Positive-only and hard plus two-sided supervision also differ in reverse targets and per-target weighting, so the comparison does not isolate the effect of a hard mask.

\section*{Limitations}

\textbf{Numerical closure.}
Section~\ref{sec:smnc-properties} analyzes an exact soft scan and hard limits.
Linear training uses shifted exponential \texttt{float32} GEMMs that can underflow cycle logits (Appendix~\ref{app:dyadic_scan_unpack}), and nonlinear training uses a clamped GEMM with artificial path evidence on acyclic supports (Appendix~\ref{app:nonlinear_induced_g}). The nonlinear barrier guard can accept points outside the M-matrix domain (Appendix~\ref{app:baseline-protocol}). The reported benchmarks were produced with these heads; exact gradient and score lower bounds should not be interpreted as guarantees on the training dynamics without a validated stable implementation.

SMNC's gradient-scaling results are local active-candidate statements (Section~\ref{sec:smnc-properties}): the \(\Theta(1/L)\) factor holds only at the minimal covering scale, while the final-scale readout \(\Omega_{\mathrm{final}}=\{K\}\) used in our cue-free linear runs (like the hard all-scale readout, Lemma~\ref{lem:coarsest-scale}) weights a length-\(L\) cycle by the length-independent factor \(2^{-K}=\Theta(1/d)\) rather than \(2^{-k_L}\); our nonlinear runs use the soft all-scale readout (\(\tau_s=0.1\)), which mixes these factors across scales.

Empirical gains are demonstrated primarily on synthetic Gaussian linear and nonlinear SEM benchmarks, with main-text summaries emphasizing RAW evaluation and TOPK ranking diagnostics reported in the appendix. As in most observational DAG-learning work, causal interpretation depends on assumptions such as causal sufficiency, faithfulness or model-specific identifiability, and appropriate benchmark design~\cite{spirtes2000causation,peters2017elements,heinze2018causal,reisach2021beware}. Soft positive reachability cues constrain path-level relations and generally do not identify a unique realizing edge. In the exploratory \(d=100\) nonlinear ablation, gains from soft supervision vary with graph regime, preprocessing, and output threshold (Appendix~\ref{app:hard-mask}). Their benefit may also depend on cue accuracy. The Sachs cues were filtered with the reference DAG. The bounded WG + SMNC pipeline introduces additional hyperparameters (including \(\beta\), \(\tau_g\), and closure-loss weights); broader robustness to graph families, sampling regimes, and hyperparameter choices remains open.

The dyadic closure scan needs \(K=\lceil\log_2 d\rceil\) dense \(d\times d\) products, so as \(d\) grows it costs more time and memory than a power-iteration spectral penalty. Non-closure penalties such as SDCD's are cheaper per step; our unstandardized benchmarks, which depart from SDCD's z-scoring protocol (Appendix~\ref{app:baseline-protocol}), do not test that regime (comparison in Appendix~\ref{app:nonwalk}).

\section*{Impact Statement}

SMNC is a scale-normalized multi-scale relation head designed to reduce product-style length bias in closure logits. Like other causal discovery methods, it produces hypotheses: inferred graphs should be combined with domain expertise, robustness checks and, when possible, interventional validation before high-stakes use.
\clearpage

\begin{ack}
This material is based upon work supported by the National Science Foundation under Grant No.~2450946. Any opinions, findings, and conclusions or recommendations expressed in this material are those of the authors and do not necessarily reflect the views of the National Science Foundation. This research is being performed using funding received from the U.S. Department of Energy Office of Nuclear Energy's Nuclear Energy University Programs under Award Number DE-NE0009383. The views and opinions of the authors expressed herein do not necessarily state or reflect those of the United States Government or any agency thereof.

The authors declare no competing interests.
\end{ack}

\newpage
\appendix
\onecolumn

\section{Additional mechanistic probes}
\label{app:mechan_more}

The probes in Appendices~\ref{app:long_cycle_decay}, \ref{app:walk_accounting} and~\ref{app:dominant_cycle_masking} are fixed-point local-sensitivity diagnostics, which evaluate gradients at a fixed input without training.

\subsection{Symmetric long-cycle sensitivity probe}
\label{app:long_cycle_decay}

Fix a directed \(L\)-cycle on \(d=L\) nodes with \(X_L(\rho)=\rho B_L\), \(\rho=0.6\), and no optimization.
We report \(q(L)=\bigl|\sum_{e\in C_L}\nabla_e\bigr|\) and per-edge \(s(L)=q(L)/L\) (symmetric cycles).
Walk-sum baselines (DAGMA log-det, NOTEARS exponential, DAG-GNN polynomial) use \(\nabla_e=\partial h/\partial A_e\) at \(A=X_L(\rho)\).
Bounded WG+SMNC maps the SMNC cycle loss on \(G=X_L(\rho)\) (\(\mu_{\mathrm{cycle}}=1\), \(\tau_g=1\)) plus the DAGMA barrier through \(A=W_f\circ G\) to an induced displacement on \(A\) (first-order chain rule in \(\log g\), \(\tau_g=1\)).
Interpretation of the curves is in Section~\ref{sec:mechan}; values below \(10^{-16}\) are plotted at that floor and log--log slopes omit exact zeros.

\subsection{Walk accounting for walk-based penalties}
\label{app:walk_accounting}

Penalties \(h(A)=\sum_{p\ge1}c_p\operatorname{tr}(B^p)\), \(B=A\circ A\), with DAGMA \(c_p=1/(p\,s^p)\), NOTEARS \(c_p=1/p!\), and DAG-GNN \(c_p=\binom dp d^{-p}\) for \(p\le d\)~\cite{dag_1,nofears}.
On an isolated directed \(L\)-cycle with edge strength \(\rho\), only multiples \(p=kL\) contribute, so each repeat traversal \(k\ge2\) adds another term.
For the exponential and polynomial penalties at typical linear-SEM weights, repeat traversals are negligible and length enters mainly through the coefficient schedule (Figure~\ref{fig:c1a_total_main}).
For DAGMA's log-det barrier, repeat traversals can dominate near the spectral-radius boundary.
These are local properties of the scalar penalties and do not by themselves predict benchmark accuracy.


\subsection{Controlled decycling under the log-det barrier and closure-side cycle loss}
\label{app:controlled-long-cycle}

The fixed-point probe in Appendix~\ref{app:long_cycle_decay} measures local
sensitivity, but not whether cycle pressure overcomes an edge-retention cost.
Here we separate reference-gradient calibration, single-cycle optimization,
two-cycle optimization, and barrier continuation.

\paragraph{Shared construction and objectives.}
On a candidate support $\mathcal E$, all edge logits are optimized independently,
all other edges are exactly masked out, and $W_{f,e}=1$, so $A=G$.
Write $g_e=G_e$, let $G^{(0)}$ be the initial gate matrix, and set
$K=\lceil\log_2d\rceil$ and $b=2^K$.
The fidelity term penalizes deviations from the initial graph:
\[
\mathcal L_{\mathrm{fid}}(G)
=\frac{\lambda_{\mathrm{fid}}}{2}
 \sum_{e\in\mathcal E}(g_e-g_e^{(0)})^2.
\]
The raw and SMNC heads use the same closure states but different final logits:
\[
\begin{aligned}
\mathcal L_{\mathrm{cyc}}^{\mathrm{raw}}(G)
&=\frac1d\sum_{i=1}^d\operatorname{softplus}(U_{ii}^{(K)}),\\
\mathcal L_{\mathrm{cyc}}(G)
&=\frac1d\sum_{i=1}^d\operatorname{softplus}(U_{ii}^{(K)}/b).
\end{aligned}
\]
Thus the diagonal average includes all $d$ nodes, whereas fidelity is a sum
over candidate edges. We compare
\[
\begin{aligned}
\text{log-det only:}\quad
&\mathcal L_{\mathrm{fid}}+h_s(A),\\
\text{log-det + raw:}\quad
&\mathcal L_{\mathrm{fid}}+h_s(A)
 +\mu_{\mathrm{cycle}}\mathcal L_{\mathrm{cyc}}^{\mathrm{raw}},\\
\text{log-det + SMNC:}\quad
&\mathcal L_{\mathrm{fid}}+h_s(A)
 +\mu_{\mathrm{cycle}}\mathcal L_{\mathrm{cyc}}.
\end{aligned}
\]
Unless stated otherwise, $\lambda_{\mathrm{fid}}=5$, $s=1$, and we use 500
full-gradient SGD updates with learning rate $0.2$, float64 arithmetic, and
the exact soft max-plus backend. A gate is removed for evaluation only when
$g_e<\delta=0.1$. The log-det-only objective is a gate-parameterized control,
not a run of the complete DAGMA learning algorithm. Unless noted otherwise
(the two-cycle runs below disable it), edge logits are clipped to
$[-30,30]$ in these single-cycle runs.

\paragraph{Reference-gradient calibration.}
For the single-cycle experiments, fix $d=32$ and embed
\[
1\xrightarrow{\rho}2\xrightarrow{\rho}\cdots
 \xrightarrow{\rho}L\xrightarrow{q_0}1,
\qquad L\in\{4,8,16,24,32\},\quad \rho=0.8,\quad q_0=0.2.
\]
The arrows give the initial gate values. Denote the closing
edge by $e_q$ and its logit by $z_q$.
To separate reference amplitude from length response, calibrate only at the
initial $L=4$ graph:
\begin{equation}
10\left|\frac{\partial\mathcal L_{\mathrm{cyc}}^{\mathrm{raw}}}{\partial z_q}\right|_{L=4}
=
\mu_{\mathrm{cycle}}^{\mathrm{match}}
\left|\frac{\partial\mathcal L_{\mathrm{cyc}}}{\partial z_q}\right|_{L=4},
\qquad \mu_{\mathrm{cycle}}^{\mathrm{match}}=89.2416.
\label{eq:controlled-cycle-calibration}
\end{equation}
Keep these coefficients fixed across $L$ and evaluate, without optimization,
\[
\begin{aligned}
g_{\mathrm{ld}}(L)
&=\left|\frac{\partial h_1(A)}{\partial z_q}\right|,\\
g_{\mathrm{raw}}(L)
&=\left|\frac{\partial[h_1(A)+10\mathcal L_{\mathrm{cyc}}^{\mathrm{raw}}]}{\partial z_q}\right|,\\
g_{\mathrm{SMNC}}^{\mathrm{match}}(L)
&=\left|\frac{\partial[h_1(A)+89.2416\mathcal L_{\mathrm{cyc}}]}{\partial z_q}\right|.
\end{aligned}
\]
Fidelity has zero gradient at initialization, and the two cycle terms have
the same sign as log-det here; hence the joint gradients also match at $L=4$.
Table~\ref{tab:controlled-cycle-gradients} shows a stronger long-cycle
response after this calibration. These are aggregate edge gradients:
the designated edge contributes to all $L$ cyclic diagonal losses.

\begin{table}[htbp]
\centering
\small
\setlength{\tabcolsep}{7pt}
\caption{Initial $z_q$ gradient at $d=32$, $s=1$. Raw uses weight $10$;
SMNC uses the reference-matched weight $89.2416$. This calibration is used
only for the gradient diagnostic.}
\label{tab:controlled-cycle-gradients}
\begin{tabular}{@{}rcccc@{}}
\toprule
$L$ & $g_{\mathrm{ld}}$ & $g_{\mathrm{raw}}$
& $g_{\mathrm{SMNC}}^{\mathrm{match}}$
& $g_{\mathrm{SMNC}}^{\mathrm{match}}/g_{\mathrm{raw}}$\\
\midrule
4  & $1.696\times10^{-2}$ & $1.524\times10^{-1}$ & $1.524\times10^{-1}$ & 1.000\\
8  & $2.820\times10^{-3}$ & $1.783\times10^{-1}$ & $2.715\times10^{-1}$ & 1.523\\
16 & $7.923\times10^{-5}$ & $9.682\times10^{-2}$ & $5.257\times10^{-1}$ & 5.429\\
24 & $2.230\times10^{-6}$ & $2.213\times10^{-2}$ & $7.636\times10^{-1}$ & 34.513\\
32 & $6.277\times10^{-8}$ & $1.584\times10^{-3}$ & $9.678\times10^{-1}$ & 610.845\\
\bottomrule
\end{tabular}
\end{table}

\begin{figure}[htbp]
\centering
\includegraphics[width=0.6\linewidth]{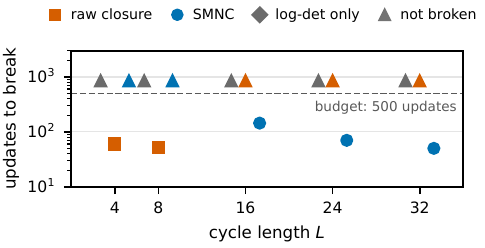}
\caption{First update at which the single cycle of
Table~\ref{tab:controlled-cycle-fixed-s} is broken (a gate falls below
$\delta=0.1$), with the same $\mu_{\mathrm{cycle}}=10$ for both closure heads.
Triangles mark runs not broken within the 500 updates.}
\label{fig:time-to-break}
\end{figure}

\paragraph{Single-cycle optimization with equal coefficients.}
We now optimize the same initial graphs with $\mu_{\mathrm{cycle}}=10$ for
\emph{both} closure heads, without reference matching.
Table~\ref{tab:controlled-cycle-fixed-s} shows that raw breaks
the $L=4,8$ cycles but SMNC does not, whereas SMNC breaks the $L=16,24,32$
cycles and raw does not; Figure~\ref{fig:time-to-break} gives the first update
at which each cycle breaks. For a cycle $C_r$ with one designated closing edge
$e_{q,r}$, path-gate retention after $N$ updates is
\[
R_r^{(N)}=
\frac{\sum_{e\in C_r\setminus\{e_{q,r}\}}g_e^{(N)}}{\rho(L_r-1)}.
\]
It is the path-gate mass after $N$ updates as a fraction of its initial value.

\begin{table}[htbp]
\centering
\small
\setlength{\tabcolsep}{8pt}
\caption{Single-cycle optimization at $d=32$, $s=1$, and 500 updates.
Both closure heads use $\mu_{\mathrm{cycle}}=10$.
Entries are final closing-edge gates; bold values are below $0.1$.
The final column is SMNC path-gate retention.}
\label{tab:controlled-cycle-fixed-s}
\begin{tabular}{@{}rcccc@{}}
\toprule
$L$ & Log-det only & + raw & + SMNC & $R^{(500)}$ (\%)\\
\midrule
4  & 0.1814 & $\mathbf{0.03918}$ & 0.1605 & 99.00\\
8  & 0.1966 & $\mathbf{0.02979}$ & 0.1471 & 98.77\\
16 & 0.1999 & 0.1045 & $\mathbf{0.007290}$ & 97.77\\
24 & 0.2000 & 0.1762 & $\mathbf{8.575\times10^{-5}}$ & 97.01\\
32 & 0.2000 & 0.1981 & $\mathbf{5.864\times10^{-6}}$ & 96.42\\
\bottomrule
\end{tabular}
\end{table}

\paragraph{Two-cycle optimization with one shared coefficient.}
\label{app:two-cycle-optimization}
Next, embed two node-disjoint cycles in $d=64$ nodes, with
$L_s=8$ and $L_\ell\in\{24,32,40,48\}$, retaining the same $\rho$, $q_0$,
fidelity, and fixed-$s$ barrier. Each cycle has its own closing edge.
Both heads use the same $\mu_{\mathrm{cycle}}\in\{60,80\}$ within each
comparison, and both cycles share that coefficient. We set $\tau_g=1$,
$\tau_c=0.1$, and $K=6$, disable logit clipping, and compare 500 and 5000
updates from identical initializations. Thresholding never removes or freezes
edges during training, and runs continue for their full budgets.
These are deterministic controlled runs on fixed graphs.

Table~\ref{tab:two-cycle-shared-weight} reports the first update at which
both thresholded cycles are broken and the long-cycle path retention after
5000 updates. Every successful run removes only the two designated closing
edges, and success persists through the final update. SMNC succeeds on all
four pairs at both weights; raw succeeds only at $L_\ell=24$, where it is
faster and retains more path-gate mass. Log-det alone leaves both cycles intact.

\begin{table}[htbp]
\centering
\small
\setlength{\tabcolsep}{5pt}
\caption{Two-cycle optimization at $d=64$, $L_s=8$, and $s=1$.
A dash means that both cycles have not been broken by update 5000.
$R_\ell^{(5000)}$ is long-cycle path-gate retention.
No additional path edges are removed. The log-det coefficient is fixed at
one and does not depend on $\mu_{\mathrm{cycle}}$.}
\label{tab:two-cycle-shared-weight}
\begin{tabular}{@{}rrccccc@{}}
\toprule
& & \multicolumn{3}{c}{First break of both cycles}
& \multicolumn{2}{c}{$R_\ell^{(5000)}$ (\%)}\\
\cmidrule(lr){3-5}\cmidrule(l){6-7}
$\mu_{\mathrm{cycle}}$ & $L_\ell$ & Log-det only & + raw & + SMNC & Raw & SMNC\\
\midrule
60 & 24 & -- & 65 & 296 & 99.88 & 98.12\\
60 & 32 & -- & -- & 296 & 98.69 & 98.01\\
60 & 40 & -- & -- & 296 & 99.54 & 97.96\\
60 & 48 & -- & -- & 296 & 99.91 & 97.92\\
\midrule
80 & 24 & -- & 43 & 124 & 99.93 & 98.00\\
80 & 32 & -- & -- & 124 & 98.52 & 97.93\\
80 & 40 & -- & -- & 124 & 99.42 & 97.89\\
80 & 48 & -- & -- & 124 & 99.89 & 97.87\\
\bottomrule
\end{tabular}
\end{table}

Increasing the budget from 500 to 5000 updates does not change any
success/failure outcome. SMNC's long-cycle path retention nevertheless
improves from $89.45\%$--$95.08\%$ to $97.87\%$--$98.12\%$; its short-cycle
retention reaches $98.57\%$--$98.75\%$.
Increasing the weight from 60 to 80 reduces the time to break both cycles
from 296 to 124 updates, with slightly greater remaining path-gate shrinkage.
The surviving raw long-cycle blocks at $L_\ell=32,40,48$ have maximum absolute
logit gradients below $7\times10^{-16}$ after 5000 updates, consistent with
near-stationary cancellation between fidelity and decycling terms.

Proposition~\ref{prop:cycle-pressure-bounds} gives
$\max\{B_{\mathrm{SMNC}}(8;64),B_{\mathrm{SMNC}}(L_\ell;64)\}\approx59.74$
for every tested pair, so both shared weights exceed this sufficient bound.
This links the experiments to a net-gradient condition; it neither estimates
the minimal successful weight nor establishes asymptotic scaling.

The earlier $d=64$, $(L_s,L_\ell)=(4,32)$, $\mu_{\mathrm{cycle}}=10$ pilot
leaves both cycles intact under SMNC, so the follow-up changes both short-cycle
length and weight. Moreover, changing $d$ from 32 to 64 changes both the
diagonal average and the final-scale normalizer. At fixed $d,K$, the two
blocks are separable: this probe tests a shared coefficient across lengths,
not competition between overlapping cycles.

\paragraph{Single-cycle barrier continuation.}
Return to the $d=32$ single-cycle construction and the equal closure weights
$\mu_{\mathrm{cycle}}=10$. We run
$s=1.0,0.9,0.8,0.7,0.6$ in sequence, warm-starting the logits, holding the
fidelity and cycle coefficients fixed, and taking 500 updates per stage.
Table~\ref{tab:controlled-cycle-continuation} reports the first stage whose
exit graph has a minimum-cost single-edge break.
This schedule uses up to 2500 updates, is not compute-matched to the
500-update fixed-$s$ run, and does not reproduce the complete benchmark
schedule. Reducing $s$ restores log-det pressure on long cycles, so log-det
continuation can also remove them. SMNC's advantage here is that it breaks
them already at $s=1$.

\begin{table}[htbp]
\centering
\small
\setlength{\tabcolsep}{12pt}
\caption{Single-cycle $s$-continuation at $d=32$ and
$\mu_{\mathrm{cycle}}=10$ for both closure heads.
Entries give the value of $s$ at the first successful stage;
a dash denotes no minimum-cost single-edge break through $s=0.6$.}
\label{tab:controlled-cycle-continuation}
\begin{tabular}{@{}rccc@{}}
\toprule
$L$ & Log-det only & + raw & + SMNC\\
\midrule
4  & --  & 1.0 & 0.6\\
8  & --  & 1.0 & 0.6\\
16 & 0.6 & 0.7 & 1.0\\
24 & 0.6 & 0.6 & 1.0\\
32 & 0.6 & 0.6 & 1.0\\
\bottomrule
\end{tabular}
\end{table}

\paragraph{Edge selection.}
Under fidelity, moving an initially retained edge to the threshold costs
$V_e(\delta)=\lambda_{\mathrm{fid}}(g_e^{(0)}-\delta)^2/2$.
Thus $q_0<\rho$ makes $e_q$ the unique minimum-cost break, $q_0=\rho$ makes
all cycle edges equivalent, and $q_0>\rho$ favors a path edge.
Additional $d=32$, $L=32$, fixed-$s$ runs with $q_0\in\{0.7,0.8,0.9\}$
do not cross the removal threshold at $\mu_{\mathrm{cycle}}=10$; for
$q_0=0.9$, the minimum gate lies on a $\rho$ path edge.
These runs serve only as edge-selection diagnostics.
Their detailed values and the full optimization traces are retained in the
reproduction materials. Normalization changes cycle pressure; it does not
by itself identify a causally incorrect edge.

\subsection{Unequal-length dominant-cycle masking probe}
\label{app:dominant_cycle_masking}

This two-cycle diagnostic asks whether a weaker cyclic component keeps a decycling signal when a stronger one coexists. The input is block-diagonal, \(X_{L_1,L_2}(\rho_1,\rho_2)=\operatorname{diag}(\rho_1B_{L_1},\rho_2B_{L_2})\) with directed-cycle supports \(B_L\) and \(\rho_1>\rho_2\); we fix \(L_1=8\), \(\rho_1=0.9\) and vary \(L_2\in\{7,8,9\}\) and \(\rho_2\in\{0.2,0.4,0.6,0.75,0.85\}\). No training is involved. DAGMA and NOTEARS are evaluated on \(A=X_{L_1,L_2}\), together with SDCD's spectral-radius gradient \(uv^\top/(u^\top v)\) from power iteration on \(A+10^{-6}\) run to convergence (500 steps from the all-ones vector). The bounded WG+SMNC curve sets \(W_{f,e}=1\) and \(G_e=\rho_k\) on the cycle edges of block \(k\), takes one infinitesimal structural step under the bounded WG+SMNC objective (\(\mu_{\mathrm{cycle}}=1\), \(\tau_g=1\), \(\tau_c=0.1\), \(\beta=5\), DAGMA barrier \(s=1.001\), \(d=L_1+L_2\), no data-fit term) and maps the signal back to the active \(A\)-space edges through the \(M,Z\to A\) chain rule. Its total includes the WG-mapped DAGMA barrier, and the ``SMNC term'' curve is its closure-side part alone. We report the absolute weak-cycle signal \(q_2=\bigl|\sum_{e\in C_2}\nabla_e\bigr|\), which avoids ratio artifacts from changes in the dominant-cycle signal.

DAGMA's log-determinant and NOTEARS's trace-exponential are additive across the two blocks: zeroing \(C_1\) leaves their signal on \(C_2\) unchanged (bit-identical for DAGMA, equal to floating-point precision for NOTEARS), so their small \(q_2\) at low \(\rho_2\) is the weak cycle's own magnitude decay, empirically \(\rho_2^{2L_2-1}\). For this block-diagonal input with \(\rho_1>\rho_2\), the spectral radius is attained by \(C_1\) alone, so the exact spectral-radius gradient vanishes on \(C_2\) even though it is cyclic (the plotted SDCD values, \(10^{-10}\) to \(3\times10^{-8}\), are the floor set by the \(10^{-6}\) shift). This is a statement about the edges of a nondominant component when one component strictly attains the spectral radius: a cycle sharing a node with \(C_1\) would receive some signal, and a tie between components would make the radius nondifferentiable. The SMNC cycle loss uses diagonal self-reachability scores, so each cyclic component contributes its own signal.

\begin{figure}[t]
  \centering
  \includegraphics[width=0.99\linewidth]{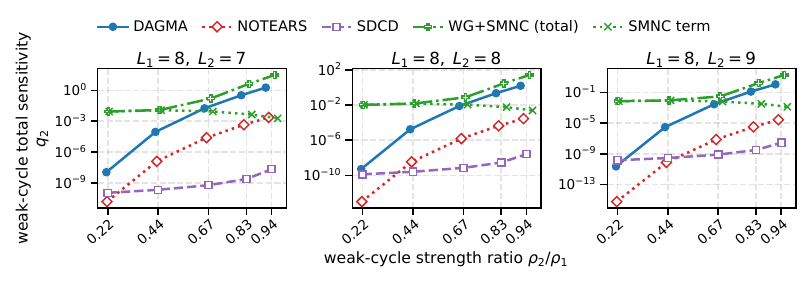}
  \caption{\textbf{Unequal-length dominant-cycle masking: weak-cycle total sensitivity \(q_2\)} against \(\rho_2/\rho_1\), for \(L_1=8\), \(\rho_1=0.9\) and \(L_2\in\{7,8,9\}\). The SMNC cycle term (dotted) gives the weaker cycle \(0.0014\)--\(0.014\) at \(\mu_{\mathrm{cycle}}=1\), more when the cycle is weak (\(0.0075\)--\(0.0101\) at \(\rho_2=0.2\)) than when it is strong (\(0.0014\)--\(0.0023\) at \(\rho_2=0.85\)); the bounded WG+SMNC total adds the WG-mapped DAGMA barrier, which dominates at large \(\rho_2\). DAGMA and NOTEARS lose weak-cycle signal through the weak cycle's own magnitude decay. SDCD's exact gradient on \(C_2\) is zero for every \(\rho_2<\rho_1\).}
  \label{fig:c1a_multilength_weak_total}
\end{figure}

The key regime is small \(\rho_2\): \(C_2\) is still a directed cycle, but its local evidence is much weaker than \(C_1\)'s. There the SMNC cycle term still gives \(C_2\) a signal, DAGMA and NOTEARS lose it to magnitude decay, and SDCD's spectral gradient is confined to \(C_1\). At large \(\rho_2\) the magnitude decay is milder, and DAGMA and NOTEARS put visible pressure on \(C_2\). A spectral penalty reaches \(C_2\) once \(C_1\)'s spectral radius falls below \(C_2\)'s; in the two-cycle optimization probe this delays the weaker cycle's removal and, at low weights, leaves it unbroken within the budget (Appendix~\ref{app:nonwalk}).

\subsection{Induced structural matrix in the nonlinear solver}
\label{app:nonlinear_induced_g}

Following the nonlinear DAGMA parameterization, we define the induced nonnegative edge-strength surrogate \(A^{\mathrm{mag,nl}}_{ij}:=\bigl(\sum_{h=1}^{m_1}(\Theta^{(1)}_{j,h,i})^{2}\bigr)^{1/2}\),
the \(\ell_2\) norm of the first-layer weights \(\Theta^{(1)}\) linking source node \(i\) to the local hidden units of target node \(j\); the implementation adds \(10^{-12}\) inside the square root. When relation-side supervision is enabled, we map it to a soft structural matrix \(G^{\mathrm{nl}}_{ij}=a/(a+c)\) with \(a=A^{\mathrm{mag,nl}}_{ij}\), \(c>0\), for \(i\neq j\) and \(G^{\mathrm{nl}}_{ii}=0\). This ratio map preserves zeros and increases monotonically toward one. The SMNC relation head and the associated cycle and cue-relation losses are applied to \(G^{\mathrm{nl}}\) with the same readout equations.

\paragraph{Clamped GEMM in the nonlinear head.}
The nonlinear implementation evaluates the soft max-plus product with an older GEMM that clamps \(X/\tau_c\) and \(Y/\tau_c\) to \([-60,60]\) before exponentiating and floors the product at \(10^{-30}\). Every composed entry is therefore at least \(\tau_c\log10^{-30}\approx-6.91\) (\(\tau_c=0.1\)), including entries with no path. For the empty graph the exact scan gives \(U_{vv}=-\infty\) and zero loss, whereas this head gives an all-scale diagonal logit of about \(0.007\) and a cycle BCE of about \(0.70\) per node (\(d=100\), \(K=7\), \(\tau_c=\tau_s=0.1\)). Compositions below the floor have zero gradient: for an isolated cycle with gates \(0.5\) in \(d=64\), the per-edge gradient of this head equals that of the exact scan for \(L\le8\) and falls to \(3\)--\(8\%\) of it for \(L\ge16\). Theorem~\ref{thm:dyadic_semantics} and Lemma~\ref{lem:soft-excess} therefore do not hold for this head, and the nonlinear results are empirical.

\section{Comparison with non-walk-based (spectral) penalties}
\label{app:nonwalk}

The main text compares SMNC with penalties that pool weighted walks: NOTEARS, the polynomial penalties and DAGMA. SDCD~\cite{benchmark_sdcd} is a different kind of method. It constrains a spectral radius, builds neither a walk sum nor a closure, and returns one scalar. We therefore keep its comparison out of the main text and report only one probe here. SDCD's benchmark rows are in Appendix~\ref{app:overview}, and its configuration is described in Appendix~\ref{app:baseline-protocol}.

\paragraph{Probe.}
On a single directed cycle, \(\rho(A)\) is the geometric mean of the edge weights, so SDCD's penalty
does not shrink with cycle length. For cycles in different strongly connected components (the probe's two
cycles are disconnected blocks), \(\rho(A)\) is the largest per-component value and, when that maximum is
unique, its exact gradient is supported on the dominant component alone. We added SDCD's penalty to the
two-cycle probe of Appendix~\ref{app:controlled-long-cycle} (Table~\ref{tab:two-cycle-shared-weight}), with
objective \(\mathcal L_{\mathrm{fid}}+h_1(A)+\mu\,h_{\mathrm{SDCD}}(A)\) and \(A=G\), where \(h_1\) is the
probe's log-det barrier shared by all methods. \(h_{\mathrm{SDCD}}\) ports \texttt{PowerIterationGradient}
from SDCD's released code (azizilab/sdcd, commit 909feb3) with its default 15 power iterations (warm-started, so the first update uses 30) and
\(10^{-6}\) shift, and matches the original class to machine precision. The short cycle has \(L_s=8\) and
geometric mean \(0.673\); the long cycle has \(L_\ell\in\{24,32,40,48\}\) and geometric mean
\(0.755\)--\(0.777\), so it is the stronger one.

\begin{figure}[t]
  \centering
  \includegraphics[width=0.85\linewidth]{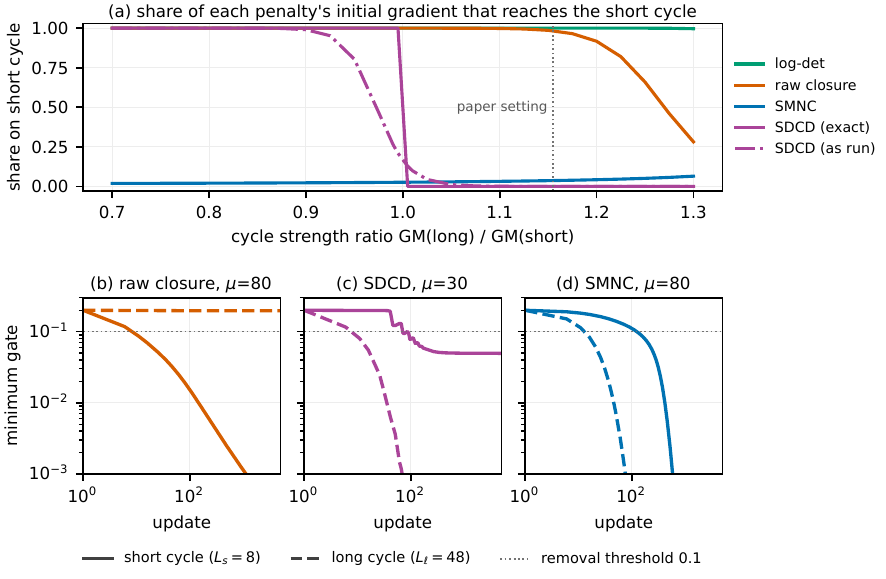}
  \caption{Probe: a short cycle (\(L_s=8\) edges) and a long cycle (\(L_\ell=48\)), disconnected from each
  other. (a) Fraction of each penalty's first-step gradient that lands on the short cycle, against how strong
  the long cycle is relative to the short one (ratio of the cycles' geometric-mean edge weights; 1 means equal,
  dotted line: the probe's setting). Exact SDCD puts all of its gradient on whichever cycle is stronger and
  switches at ratio 1; ``as run'' is SDCD's warm-started power iteration (30 iterations by the first update).
  (b)--(d) Smallest gate on each cycle during training; a cycle is removed once its smallest gate falls below
  0.1 (dotted). SDCD and SMNC remove the long cycle first and the short one later; raw closure never removes
  the long cycle.}
  \label{fig:sdcd-allocation}
  \vspace{6pt}
  \includegraphics[width=0.9\linewidth]{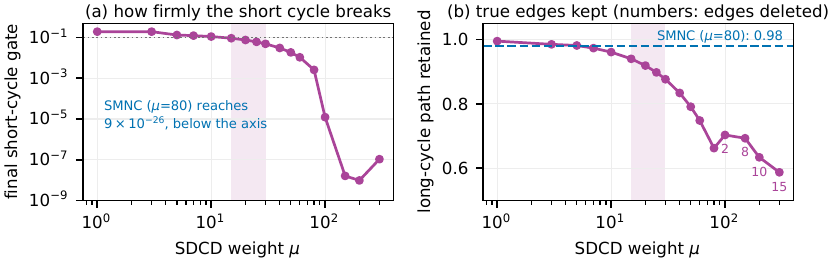}
  \caption{The same probe (\(L_\ell=48\), 5000 updates) with SDCD at a constant weight \(\mu\) (how hard the
  penalty pushes) and its default \(10^{-6}\) shift. (a) Final gate of the short cycle's closing edge; below the
  dotted line (0.1) the cycle is removed. (b) Fraction of the long cycle's path-gate mass kept; numbers give true
  path edges deleted. Shading: weights at which SDCD removes both cycles at every
  \(L_\ell\in\{24,32,40,48\}\) without deleting extra edges. SMNC at \(\mu=80\) is shown for reference
  (annotated in a, dashed in b).}
  \label{fig:sdcd-weights}
\end{figure}

\paragraph{Results.}
At the first update the exact spectral gradient on the short cycle is zero, and SDCD's power iteration as run
sends it at most \(5\times10^{-4}\) of the total (Figure~\ref{fig:sdcd-allocation}a), against
\(4\%\)--\(12\%\) for SMNC, whose share stays nonzero at every strength ratio. SMNC's split follows cycle
length, though: across the sweep the short cycle receives \(2\%\)--\(24\%\), and less when it is the stronger
cycle, consistent with SMNC also breaking it late. At \(L_\ell=48\) (Figure~\ref{fig:sdcd-allocation}b--d), SDCD
(\(\mu=30\)) breaks the long cycle at update 8 and the short one at update 70, SMNC (\(\mu=80\)) breaks them at
updates 13 and 124, and raw closure (\(\mu=80\)) breaks the short cycle at update 8 and never breaks the long
one. In every one of the more than 160 SDCD runs in which a cycle broke (all weights, schedules and
optimizers), the long, stronger cycle broke first or at the same update.

\paragraph{Sensitivity to the weight.}
With a constant \(\mu\in[15,30]\), SDCD breaks both cycles at every \(L_\ell\) without removing an extra path
edge (shaded in Figure~\ref{fig:sdcd-weights}), but the break of the short cycle is shallow (at \(L_\ell=48\)
its final gate is \(0.094\) at \(\mu=15\) and \(0.050\) at \(\mu=30\), so a removal threshold of \(0.05\) would
leave it in place for \(\mu=15\)--\(25\)), and across \(L_\ell\) SDCD retains only \(81\%\)--\(94\%\) of the
long path's gate mass, against \(98\%\) for SMNC at \(\mu=80\). Below this range the short cycle survives at
\(L_\ell=48\) (its gate stays at \(0.113\) for \(\mu=10\)); above it SDCD removes true path edges. Most of the
shallow break comes from the constant \(10^{-6}\) shift: lowering it to \(10^{-12}\) breaks both cycles at every
\(L_\ell\) for \(\mu=10\) and \(30\), with one exception to clean behavior: at \(\mu=30\), \(L_\ell=32\) it
removes one extra path edge and keeps \(96.7\%\) of the long path.

\paragraph{Limits of the comparison.}
SDCD is not weaker at breaking cycles in general. On a single isolated cycle (the \(d=32\) protocol of
Table~\ref{tab:controlled-cycle-fixed-s}, with SDCD's penalty in place of the closure head), SDCD at the
probe's common weight \(\mu=10\) (not a matched scale) breaks every length \(L\in\{4,8,16,24,32\}\) within 18
updates without extra deletions, whereas SMNC at that weight breaks only \(L\ge16\). Breaking the long cycle
first is not specific to SDCD: SMNC also breaks it first, at updates 13--36 versus 124--296 under SGD
(\(\mu=60,80\)). The weights are not on a common scale: SDCD was swept finely (\(\mu\) from 1 to 300), while raw
closure and SMNC are shown only at their archived weights (\(\mu=60, 80\)), so the width of each method's
working range has not been compared. The probe also grafts SDCD's penalty onto the gate-only controlled
problem (no MLP or sparsity term, a fidelity term in place of SDCD's likelihood, no stage-1 prescreening,
float64 arithmetic and sigmoid gates in place of SDCD's weight-norm adjacency), so it characterizes the
penalty alone.


\clearpage
\section{Proofs and additional technical details for Section~\ref{third:method}}
\label{app:method-proofs}

\begin{figure}[H]
  \centering
  \includegraphics[width=\linewidth]
    {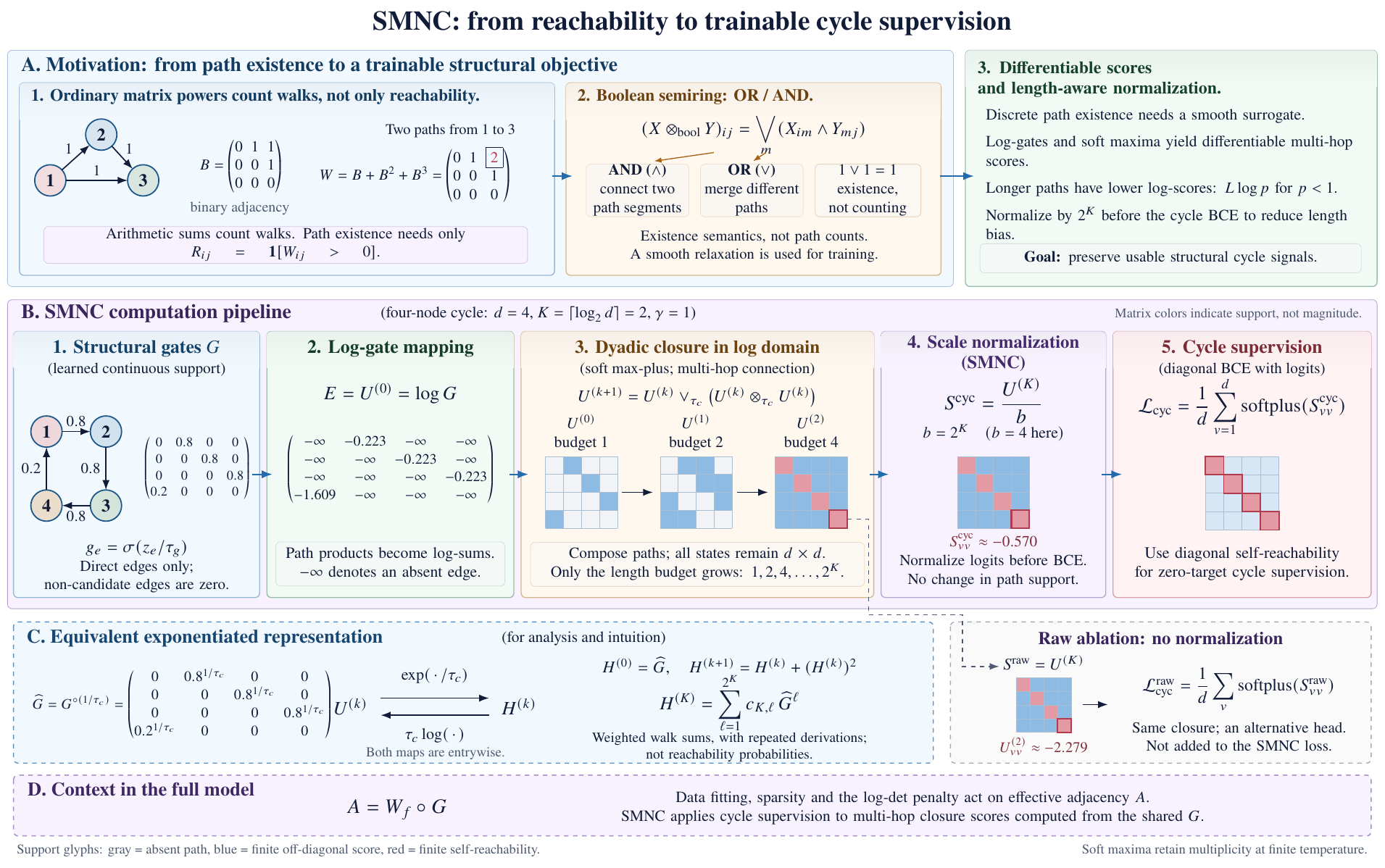}

  \caption{\textbf{Detailed SMNC computation and its exponentiated
  walk representation.}
  (A) Motivation: arithmetic matrix powers count walks, whereas
  Boolean closure encodes path existence.
  (B) A four-node cycle illustrates the continuous computation:
  structural gates are mapped to log-space, propagated through
  dyadic closure, and scale-normalized before diagonal cycle BCE.
  The matrices remain $4\times4$ while the walk-length budget
  increases from $1$ to $2$ to $4$.
  (C) The exponentiated state provides an equivalent weighted-walk
  representation; the raw ablation bypasses scale normalization.
  (D) Connection to the effective adjacency $A=W_f\circ G$.
  Matrix colors indicate path support, not score magnitude.}
  \label{fig:smnc-computation-overview}
\end{figure}

\subsection{Cue-free SMNC training: pseudocode}
\label{app:smnc-pseudocode}

Algorithm~\ref{alg:smnc} summarizes the cue-free training loop of the linear bounded-WG solver defined in
Section~\ref{sec:wg-factorized}--Section~\ref{sec:reach-regularized}: the
factorized effective adjacency \(A=(W_f(M)\circ G(Z))\circ(\mathbf 1-I)\)
(Eq.~\eqref{eq:A_WG}); the log-gate scores \(E=\log G(Z)\) with \(E_{ii}=-\infty\)
(Section~\ref{sec:wg-factorized}); the dyadic soft max-plus closure scan at
temperature \(\tau_c\) (Eq.~\eqref{eq:smnc-soft-recursion}), whose exact hard
(\(\vee,\otimes\)) semantics are given by Theorem~\ref{thm:dyadic_semantics};
scale normalization of each dyadic state by \(2^{\gamma k}\)
(Eq.~\eqref{eq:smnc-normalized-scale}); the readout on a scale set \(\Omega\)
(Eq.~\eqref{eq:smnc-readout-soft}; Section~\ref{sec:synth} states the scale set
used for each benchmark); the zero-target diagonal cycle BCE on the resulting
cycle logit \(S^{\mathrm{cyc}}\) (Section~\ref{sec:unified-bce}); and the
DAGMA-style outer loop over the continuation schedule \(s\) that optimizes the
total cue-free objective of Eq.~\eqref{eq:wg-dagma-obj} at
\(\mu_{\mathrm{rel}}=0\) (Section~\ref{sec:reach-regularized}).

\begin{algorithm}[t]
\caption{Cue-free SMNC training of the linear bounded-WG solver (cycle-regularized continuous DAG learning)}
\label{alg:smnc}
\begin{algorithmic}[1]
\Require data \(X\); outer continuation schedule \(\{s_t\}_{t=1}^{N_{\mathrm{outer}}}\); magnitude bound \(\beta\); gate temperature \(\tau_g\); closure temperature \(\tau_c\); scale-normalization exponent \(\gamma\); readout scale set \(\Omega\) and temperature \(\tau_s\) (Section~\ref{sec:synth} states \(\Omega\) per benchmark); weights \(\lambda_1,\mu_{\mathrm{cycle}}\); initial data-fit weight \(\mu_{\mathrm{init}}\) and its decay factor; number of inner steps per stage; magnitude threshold \(\omega\)
\State Initialize magnitude logits \(M\) and gate logits \(Z\); \(\mu\gets\mu_{\mathrm{init}}\)
\For{\(t=1,\dots,N_{\mathrm{outer}}\)} \Comment{DAGMA-style outer loop, Eq.~\eqref{eq:wg-dagma-obj}}
  \State \(s\gets s_t\)
  \For{each of the stage's fixed number of inner steps (Section~\ref{sec:metrics})} \Comment{warm-started from stage \(t-1\)}
    \State \(W_f(M)\gets\beta\tanh(M)\); \quad \(G(Z)\gets\sigma(Z/\tau_g)\circ(\mathbf 1-I)\) \Comment{Eq.~\eqref{eq:A_WG}}
    \State \(A\gets(W_f(M)\circ G(Z))\circ(\mathbf 1-I)\)
    \State \(U^{(0)}\gets\log G(Z)\), with diagonal set to \(-\infty\) \Comment{log-gate scores \(E\)}
    \For{\(k=0,\dots,K-1\)} \Comment{\(K=\lceil\log_2 d\rceil\)}
      \State \(U^{(k+1)}\gets U^{(k)}\vee_{\tau_c}\bigl(U^{(k)}\otimes_{\tau_c}U^{(k)}\bigr)\) \Comment{Eq.~\eqref{eq:smnc-soft-recursion}}
    \EndFor
    \For{\(k=0,\dots,K\)}
      \State \(\widetilde U^{(k)}\gets U^{(k)}/2^{\gamma k}\) \Comment{scale normalization, Eq.~\eqref{eq:smnc-normalized-scale}}
    \EndFor
    \State \(S^{(\gamma)}(\Omega)\gets\widetilde U^{(k)}\) if \(\Omega=\{k\}\), else \(\tau_s\log\sum_{k\in\Omega}\exp\!\bigl(\widetilde U^{(k)}/\tau_s\bigr)\) \Comment{readout, Eq.~\eqref{eq:smnc-readout-soft}}
    \State \(S^{\mathrm{cyc}}\gets S^{(\gamma)}(\Omega)\) \Comment{diagonal cycle logit, Section~\ref{sec:unified-bce}}
    \State \(\mathcal L_{\mathrm{cyc}}\gets\frac1d\sum_v\operatorname{softplus}\!\bigl(S^{\mathrm{cyc}}_{vv}\bigr)\) \Comment{zero-target diagonal cycle BCE}
    \State \(h_s(A)\gets d\log s-\log\det(sI-A\circ A)\) \Comment{Eq.~\eqref{eq:dagma-barrier}}
    \State \(\mathcal L\gets\mu\bigl(\mathrm{score}(A)+\lambda_1\lVert A\rVert_1\bigr)+h_s(A)+\mu_{\mathrm{cycle}}\mathcal L_{\mathrm{cyc}}\) \Comment{Eq.~\eqref{eq:wg-dagma-obj}, \(\mu_{\mathrm{rel}}=0\)}
    \State Take a gradient step on \(\mathcal L\) with respect to \((M,Z)\)
  \EndFor
  \State \(\mu\gets\mu\cdot{}\)(decay factor) \Comment{DAGMA-style decrease of the data-fit weight}
\EndFor
\State \(A\gets(W_f(M)\circ G(Z))\circ(\mathbf 1-I)\); set \(A_{ij}\gets0\) wherever \(|A_{ij}|<\omega\) \Comment{\(\omega=0.1\) on most datasets, Section~\ref{sec:metrics}}
\State \Return \(A\)
\end{algorithmic}
\end{algorithm}

\subsection{Proof of the hard dyadic scan semantics}
\label{app:proof-dyadic-semantics}

\begin{proof}[Proof of Theorem~\ref{thm:dyadic_semantics}]
Write \(W^{(k)}_{ij}\) for the right-hand side, the best score of an \(i\to j\) walk of length in \([1,2^k]\). We prove \(U^{(k)}=W^{(k)}\) by induction on \(k\). For \(k=0\), \(U^{(0)}=E\) and the only walks of length one are single edges, so \(U^{(0)}_{ij}=E_{ij}=W^{(0)}_{ij}\) (with \(E_{ii}=-\infty\) forbidding self-loops).

Assume \(U^{(k)}=W^{(k)}\). The update
\[
U^{(k+1)}_{ij}=\max\Bigl\{U^{(k)}_{ij},\ \max_m\bigl(U^{(k)}_{im}+U^{(k)}_{mj}\bigr)\Bigr\}
\]
takes the maximum of a skip term and a concatenation term.

(\(\le\)) The skip term is the score of a walk of length at most \(2^k\le2^{k+1}\). Each concatenation term \(U^{(k)}_{im}+U^{(k)}_{mj}\) is, by the induction hypothesis, the score of an \(i\to m\) walk followed by an \(m\to j\) walk, each of length in \([1,2^k]\), i.e.\ of an \(i\to j\) walk of length in \([2,2^{k+1}]\). Hence \(U^{(k+1)}_{ij}\le W^{(k+1)}_{ij}\).

(\(\ge\)) Let \(\pi\) be any \(i\to j\) walk of length \(\ell\in[1,2^{k+1}]\). If \(\ell\le 2^k\), its score is at most \(W^{(k)}_{ij}=U^{(k)}_{ij}\), the skip term. Otherwise \(2^k<\ell\le2^{k+1}\); let \(m\) be the vertex reached after the first \(2^k\) steps of \(\pi\). This splits \(\pi\) into an \(i\to m\) walk of length \(2^k\) and an \(m\to j\) walk of length \(\ell-2^k\in[1,2^k]\), whose scores are at most \(U^{(k)}_{im}\) and \(U^{(k)}_{mj}\) by the induction hypothesis. Hence the score of \(\pi\) is at most \(U^{(k)}_{im}+U^{(k)}_{mj}\le U^{(k+1)}_{ij}\). Maximizing over \(\pi\) gives \(W^{(k+1)}_{ij}\le U^{(k+1)}_{ij}\), which completes the induction.

Finally, since \(E\le 0\), every closed walk has non-positive score. If a maximizing walk repeats a vertex, deleting the closed sub-walk between the two visits does not lower its score and shortens it, so it stays within the budget. Repeating this yields a simple path if \(i\neq j\). If \(i=j\), first cut the walk at its first return to \(i\) and then delete closed sub-walks among the intermediate vertices; the result is a simple cycle through \(i\) (of length at least two, since \(E_{ii}=-\infty\)).
\end{proof}

\subsection{Proof of the length-stable cycle lower bound}
\label{app:proof-length-stable-cycle}

\begin{proof}[Proof of Theorem~\ref{thm:length_stable_cycle}]
Since \(|A_e|=|W_{f,e}|G_e\le \beta G_e\), the assumption \(|A_e|\ge \alpha\)
implies \(G_e\ge \alpha/\beta\) for every edge \(e\in C\).
For any node \(v\) on the cycle, the cycle path from
\(v\) back to \(v\) has length \(L\) and is feasible at every dyadic scale
\(a\) such that \(2^a\ge L\). Hence, for any such scale,
\[
U_{vv}^{(a)}
\ge
L\log(\alpha/\beta).
\]
Since \(\log(\alpha/\beta)\le 0\), dividing by \(2^a\) gives
\[
\frac{U_{vv}^{(a)}}{2^a}
\ge
\frac{L}{2^a}\log(\alpha/\beta).
\]
For \(a=k_L=\lceil\log_2 L\rceil\), we have \(L\le 2^{k_L}\), so
\(L/2^{k_L}\le 1\). Therefore
\[
\frac{U_{vv}^{(k_L)}}{2^{k_L}}
\ge
\log(\alpha/\beta).
\]
For \(a=K=\lceil\log_2 d\rceil\), the cycle is simple and hence \(L\le d\le
2^K\), so \(L/2^K\le 1\). Therefore
\[
\frac{U_{vv}^{(K)}}{2^K}
\ge
\log(\alpha/\beta).
\]
The smooth recursion and smooth readouts are log-sum-exp relaxations
of the corresponding hard maxima and therefore, in exact arithmetic, dominate the hard max values
(Lemma~\ref{lem:soft-excess}).
The same lower bounds follow for the corresponding smooth readouts; a finite-precision
implementation can violate them where its exponential-space products underflow
(Section~\ref{sec:smnc-properties}).
\end{proof}

\subsection{Numerically stabilized GEMM-oriented algebra}
\label{app:dyadic_scan_unpack}

In the linear solver, each soft composition is evaluated as one GEMM in shifted exponential space: with row shifts \(r_i=\max_m X_{im}\) and column shifts \(c_j=\max_m Y_{mj}\),
\((X\otimes_{\tau_c}Y)_{ij}=r_i+c_j+\tau_c\log\sum_m e^{(X_{im}-r_i)/\tau_c}e^{(Y_{mj}-c_j)/\tau_c}\), computed in \texttt{float32}. Masked entries (\(E=-\infty\)) contribute exactly zero, and entries whose product underflows are set to \(-\infty\). The shift leaves \(U^{(k)}\) unchanged in exact arithmetic, but because it is per row and column rather than per entry, in \texttt{float32} it can drop an \((i,j)\) entry whose best composition lies far below \(r_i+c_j\), e.g.\ a long weak cycle next to strong incident edges (Section~\ref{sec:smnc-properties}). Relation logits are clipped to \([-50,50]\). Products are floored at the smallest normal \texttt{float32} value before the logarithm, which also distorts very small positive products. The nonlinear solver uses a clamped variant of this GEMM (Appendix~\ref{app:nonlinear_induced_g}).
Exact-semantic theorems in Section~\ref{sec:smnc-properties} should not be read as guarantees on \texttt{float32} training dynamics without a validated stable scan.

\subsection{Proofs of the gradient factor, coarsest-scale dominance, and the soft-scan excess}
\label{app:proof-coarsest-scale}

\begin{proof}[Proof of Corollary~\ref{cor:poly_gradient_decay_bce}]
By Theorem~\ref{thm:dyadic_semantics}, \(U^{(a)}_{vv}\) is a maximum over the finite set of closed walks through
\(v\) of length at most \(2^a\). If the cycle candidate \(C\) is its unique maximizer, the maximum is still
attained only by \(C\) in a neighborhood of the current gates, and there
\(U^{(a)}_{vv}=\sum_{e\in C}\log g_e\). Because \(E<0\) off the diagonal, deleting a closed sub-walk strictly
increases a walk score, so a unique maximizer is a simple cycle and each of its edges appears once. Hence
\(\partial S_{vv,a}/\partial\log g_e=2^{-a}\), and \(\operatorname{softplus}'=\sigma\) gives
\(\partial\ell^{\rm cyc}_{v,a}/\partial\log g_e=2^{-a}\sigma(S_{vv,a})\). Finally,
\(2^{k_L-1}<L\le2^{k_L}\) gives \(2^{-k_L}\in(1/(2L),1/L]\), and \(2^{K-1}<d\le2^K\) gives
\(2^{-K}\in(1/(2d),1/d]\).
\end{proof}

\begin{lemma}[Coarsest-scale dominance of the hard readout]
\label{lem:coarsest-scale}
For the hard scan and any \(\gamma>0\), \(U^{(a)}_{ij}/2^{\gamma a}\) is non-decreasing in \(a\). Hence
\(S^{(\gamma),\max}_{ij}(\Omega)=U^{(\max\Omega)}_{ij}/2^{\gamma\max\Omega}\), and the hard
\(\Omega_{\mathrm{all}}\) and \(\Omega_{\mathrm{final}}\) readouts coincide.
\end{lemma}

\begin{proof}[Proof of Lemma~\ref{lem:coarsest-scale}]
The skip term in Eq.~\eqref{eq:smnc-hard-recursion} gives \(U^{(a+1)}\ge U^{(a)}\) entrywise, and
\(U^{(a)}\le 0\) because every walk score is a sum of entries \(E_e\le 0\). Hence, for any \(\gamma>0\),
\[
\frac{U^{(a+1)}_{ij}}{2^{\gamma(a+1)}}\ \ge\ \frac{U^{(a)}_{ij}}{2^{\gamma(a+1)}}\ \ge\ \frac{U^{(a)}_{ij}}{2^{\gamma a}},
\]
where the second step holds because dividing a non-positive number by a larger positive number cannot
decrease it (the chain is read in the extended-real order and is immediate when \(U^{(a)}_{ij}=-\infty\)). Thus the maximum over \(\Omega\) is
attained at \(\max\Omega\). If \(U^{(a)}_{ij}\) is finite it is strictly negative (off-diagonal gates lie in
\((0,1)\)), the second inequality is strict, and the active scale \(\max\Omega\) is unique.
\end{proof}

\paragraph{The bound in ``Objective used in our experiments''.}
Let \(C\) satisfy the assumptions of Theorem~\ref{thm:length_stable_cycle} and be the unique maximizer of
\(U^{(K)}_{vv}\) for every \(v\in C\). By Corollary~\ref{cor:poly_gradient_decay_bce} with \(a=K\), the diagonal
terms of the nodes on \(C\) contribute
\(\frac{1}{d\,2^K}\sum_{v\in C}\sigma(U^{(K)}_{vv}/2^K)\) to \(\partial\mathcal L_{\mathrm{cyc}}/\partial\log g_e\)
for every \(e\in C\). Theorem~\ref{thm:length_stable_cycle} gives \(U^{(K)}_{vv}/2^K\ge\log(\alpha/\beta)\), and
\(U^{(K)}_{vv}\le0\) gives \(\sigma(U^{(K)}_{vv}/2^K)\le1/2\); with \(d\le2^K<2d\) the contribution lies in
\([L\sigma(\log(\alpha/\beta))/(2d^2),\,L/(2d^2)]\). For the unnormalized head \(\Lambda=U^{(K)}\), the same terms
contribute \(\frac1d\sum_{v\in C}\sigma(U^{(K)}_{vv})\), and \(U^{(K)}_{vv}=\sum_{e\in C}\log g_e\le L\log p_{\max}\).
Diagonal terms of other nodes add a non-negative amount wherever the hard readout is differentiable, since
\(U^{(K)}\) is non-decreasing in every \(\log g_e\).
These are derivatives in log-gate coordinates at fixed \(d\), for the hard scan. With \(g_e=\sigma(Z_e/\tau_g)\), the derivative with respect to the trained parameter carries the extra factor \((1-g_e)/\tau_g\), which vanishes as \(g_e\to1\). At finite \(\tau_c\), soft values dominate hard values (Lemma~\ref{lem:soft-excess}), but this does not imply a corresponding lower bound on the per-edge gradient.

\begin{lemma}[Walk expansion and excess of the soft scan]
\label{lem:soft-excess}
Let \(U^{(k)}_{\mathrm{soft}}\) follow Eq.~\eqref{eq:smnc-soft-recursion} and \(U^{(k)}_{\mathrm{hard}}\)
follow Eq.~\eqref{eq:smnc-hard-recursion}, from the same \(E\), and let \(T:=\exp(E/\tau_c)=G^{\circ1/\tau_c}\)
(zero diagonal).
\par (i) \(\exp(U^{(k)}_{\mathrm{soft}}/\tau_c)=\sum_{\ell=1}^{2^k}c_{k,\ell}T^\ell\) with integer coefficients
\(c_{k,\ell}\ge1\) that do not depend on \(G\).
\par (ii) \(0\le U^{(k)}_{\mathrm{soft},ij}-U^{(k)}_{\mathrm{hard},ij}\le(2^k-1)\tau_c\log(2d)\) whenever \(U^{(k)}_{\mathrm{hard},ij}>-\infty\);
both states are \(-\infty\) at exactly the same entries.
\par (iii) For \(\gamma=1\) and \(a\le b\),
\(\widetilde U^{(a)}_{\mathrm{soft},ij}\le\widetilde U^{(b)}_{\mathrm{soft},ij}+\tau_c\log(2d)\), and the smooth
readout of Eq.~\eqref{eq:smnc-readout-soft} satisfies
\(S^{(1)}_{ij}(\Omega_{\mathrm{final}})\le S^{(1)}_{ij}(\Omega_{\mathrm{all}})\le
S^{(1)}_{ij}(\Omega_{\mathrm{final}})+\tau_c\log(2d)+\tau_s\log(K+1)\).
\end{lemma}
\begin{proof}
(i) Let \(H^{(k)}:=\exp(U^{(k)}_{\mathrm{soft}}/\tau_c)\). Since \(\exp((a\vee_{\tau_c}b)/\tau_c)=e^{a/\tau_c}+e^{b/\tau_c}\)
and \(\exp((X\otimes_{\tau_c}Y)_{ij}/\tau_c)=\sum_m e^{X_{im}/\tau_c}e^{Y_{mj}/\tau_c}\), the soft recursion is
\(H^{(k+1)}=H^{(k)}+(H^{(k)})^2\) with ordinary matrix multiplication and \(H^{(0)}=T\). If
\(H^{(k)}=\sum_{\ell=1}^{2^k}c_{k,\ell}T^\ell\) with integers \(c_{k,\ell}\ge1\), then \(H^{(k+1)}\) has integer
coefficients; for \(\ell\le2^k\) they are at least \(c_{k,\ell}\ge1\), and for \(2^k<\ell\le2^{k+1}\) they are
\(\sum_{\ell_1+\ell_2=\ell}c_{k,\ell_1}c_{k,\ell_2}\ge1\), since \(\ell=2^k+(\ell-2^k)\) with both parts in \([1,2^k]\).

(ii) Both soft operations are monotone in their arguments and dominate the corresponding maxima, so
\(U_{\mathrm{soft}}^{(k)}\ge U_{\mathrm{hard}}^{(k)}\) by induction; an entry is \(-\infty\) in either scan exactly
when all of its terms are. For the upper bound, log-sum-exp over \(n\) terms exceeds their maximum by at most
\(\tau_c\log n\), and \(\otimes_{\tau_c}\) has \(d\) terms while \(\vee_{\tau_c}\) has two. Let \(g_k\ge0\) bound
\(U^{(k)}_{\mathrm{soft}}-U^{(k)}_{\mathrm{hard}}\) on finite entries. Then
\((U^{(k)}_{\mathrm{soft}}\otimes_{\tau_c}U^{(k)}_{\mathrm{soft}})_{ij}\le(U^{(k)}_{\mathrm{hard}}\otimes U^{(k)}_{\mathrm{hard}})_{ij}+2g_k+\tau_c\log d\),
so \(U^{(k+1)}_{\mathrm{soft},ij}\le U^{(k+1)}_{\mathrm{hard},ij}+2g_k+\tau_c\log(2d)\). With \(g_0=0\), the recursion
\(g_{k+1}=2g_k+\tau_c\log(2d)\) gives \(g_k=(2^k-1)\tau_c\log(2d)\).

(iii) By (ii) and Lemma~\ref{lem:coarsest-scale},
\(\widetilde U^{(a)}_{\mathrm{soft}}\le\widetilde U^{(a)}_{\mathrm{hard}}+(1-2^{-a})\tau_c\log(2d)\le
\widetilde U^{(b)}_{\mathrm{hard}}+\tau_c\log(2d)\le\widetilde U^{(b)}_{\mathrm{soft}}+\tau_c\log(2d)\). The
readout over the \(K+1\) scales of \(\Omega_{\mathrm{all}}\) lies between its largest term, which is at least
\(\widetilde U^{(K)}_{\mathrm{soft}}=S^{(1)}(\Omega_{\mathrm{final}})\), and that term plus \(\tau_s\log(K+1)\);
applying the first claim with \(b=K\) completes the proof.
\end{proof}
Thus, in contrast to the hard scan, soft states can be positive.
Off-diagonal normalized soft scores need not be monotone in \(a\) (Lemma~\ref{lem:soft-excess}(iii)).
For diagonals in the exact soft recursion, \(H^{(k+1)}_{vv}\ge(H^{(k)}_{vv})^2\), so at \(\gamma=1\),
\(\widetilde U_{\mathrm{soft},vv}^{(k+1)}\ge\widetilde U_{\mathrm{soft},vv}^{(k)}\).
Part (i) is the
tempered walk-series form used in Section~\ref{sec:wg-factorized}: each extra traversal of a cycle with gate
product \(Q\) multiplies a walk term by \(Q^{1/\tau_c}\), but the fixed counts \(c_{k,\ell}\) grow rapidly with \(\ell\) over most of the range (e.g.\ \(c_{7,\ell}=1,7,42,231,\dots\), peaking near \(10^{25}\) at \(\ell=74\)),
so repeated traversals are discounted, not removed. On an isolated cycle the multiplicity enters the gradient
through \(r=\partial U^{(K)}_{vv}/\partial\log Q=\sum_m m\,\omega_m\ge1\)
(Appendix~\ref{app:cycle-weight-bounds}); for \(d=128\) and \(\tau_c=0.1\), a 2-cycle with both gates \(0.9\) has
\(r\approx11.7\) and one with gates \(0.7\) has \(r\approx1.03\).

\begin{remark}[Active scale and length behavior]
\label{rem:old_vs_new_decay}
On a uniform-gate chain or cycle, a raw product-style path log-score scales as \(L\log p\), so a zero-target BCE factor behaves like \(\sigma(L\log p)\approx p^L\). SMNC instead normalizes dyadic path scores by their scale budgets. For the hard all-scale readout, \(S_{vv}^{(1),\max}(\Omega_{\mathrm{all}})=\max_{a=0,\ldots,K}U^{(a)}_{vv}/2^a\), and by Lemma~\ref{lem:coarsest-scale} the active scale is \(a^\star=K=\lceil\log_2 d\rceil\) whenever \(U^{(K)}_{vv}\) is finite, so the local factor is \(2^{-K}=\Theta(1/d)\), the same as for \(\Omega_{\mathrm{final}}\) (the readout of our linear runs). The \(\Theta(1/L)\) factor of Corollary~\ref{cor:poly_gradient_decay_bce} is a minimal-covering-scale statement, not a property of the all-scale or final-scale readouts. With smooth scale aggregation (\(\tau_s>0\), as in our nonlinear runs) and the hard scan, \(\partial S^{(1)}_{vv}(\Omega_{\mathrm{all}})/\partial\log g_e=\sum_a w_a\,\partial\widetilde U^{(a)}_{vv}/\partial\log g_e\) with \(w_a\propto\exp(\widetilde U^{(a)}_{vv}/\tau_s)\). Because \(U^{(a)}_{vv}\le U^{(K)}_{vv}\le0\), \(w_a/w_K\le\exp\bigl(-(2^{K-a}-1)\lvert\widetilde U^{(K)}_{vv}\rvert/\tau_s\bigr)\), so finer scales contribute only when \(\lvert\widetilde U^{(K)}_{vv}\rvert\lesssim\tau_s\). On an isolated length-\(L\) cycle the factor is \(\sum_{a\ge k_L}w_a2^{-a}\in[2^{-K},2^{-k_L}]\); for example, at \(d=128\), \(\tau_s=0.1\), \(L=3\) and gates \(0.95\), the per-edge BCE derivative is about \(3.9\times10^{-2}\), versus \(3.9\times10^{-3}\) for \(\Omega_{\mathrm{final}}\).
\end{remark}
\subsection{Sufficient cycle-weight bounds for the controlled probes}
\label{app:cycle-weight-bounds}

This subsection analyzes the fidelity--cycle competition in
Appendix~\ref{app:controlled-long-cycle}. The statement concerns isolated
cycle supports only. Write $\mathcal L_{\mathrm{raw}}=
\mathcal L_{\mathrm{cyc}}^{\mathrm{raw}}$ and
$\mathcal L_{\mathrm{SMNC}}=\mathcal L_{\mathrm{cyc}}$ as defined there.

\begin{proposition}[Sufficient pressure on an intact isolated cycle]
\label{prop:cycle-pressure-bounds}
Let the allowed support be a union of node-disjoint simple directed cycles
and isolated nodes. On every allowed edge, let
$g_e=\sigma(z_e/\tau_g)$ with finite $z_e$ and $\tau_g>0$, and set $A=G$.
Use the exact masked recursion in Eq.~\eqref{eq:smnc-soft-recursion}, with
one fixed temperature $\tau_c>0$ for both soft operations, no zero-length
walks, and no logit clipping. Let $K=\lceil\log_2d\rceil$ and $b=2^K$.
Consider
\[
\mathcal J_{\mathrm{raw}}=\mathcal L_{\mathrm{fid}}+\nu h_1(G)
 +\mu_{\mathrm{cycle}}\mathcal L_{\mathrm{raw}},\qquad
\mathcal J_{\mathrm{SMNC}}=\mathcal L_{\mathrm{fid}}+\nu h_1(G)
 +\mu_{\mathrm{cycle}}\mathcal L_{\mathrm{SMNC}},
\]
where $\lambda_{\mathrm{fid}}>0$, $\nu\ge0$, and
$\mu_{\mathrm{cycle}}\ge0$. For one cycle $C_L$, let its designated edge have
initial gate $q_0$ and current gate $q$, with $0<\delta<q_0<1$.
Define
\begin{equation}
\begin{aligned}
B_{\mathrm{raw}}(L;d)
&:=\frac{\lambda_{\mathrm{fid}}q_0^2d}{4L}
       (1+\delta^{-L}),\\
B_{\mathrm{SMNC}}(L;d)
&:=\frac{\lambda_{\mathrm{fid}}q_0^2db}{4L}
       (1+\delta^{-L/b}),\\
B_{\mathrm{ld}}(L)
&:=\frac{\lambda_{\mathrm{fid}}q_0^2}{8}
       (\delta^{-2L}-1).
\end{aligned}
\label{eq:cycle-sufficient-weight-bounds}
\end{equation}
At any state where all edges of $C_L$ satisfy $g_e\ge\delta$,
$\mu_{\mathrm{cycle}}>B_{\mathrm{raw}}$ guarantees
$\partial\mathcal J_{\mathrm{raw}}/\partial z_q>0$, and
$\mu_{\mathrm{cycle}}>B_{\mathrm{SMNC}}$ guarantees
$\partial\mathcal J_{\mathrm{SMNC}}/\partial z_q>0$.
For the log-det-only objective
$\mathcal J_{\mathrm{ld}}=\mathcal L_{\mathrm{fid}}+\nu h_1(G)$,
$\nu>B_{\mathrm{ld}}$ gives the analogous guarantee.
For several cycles, a common weight exceeding the maximum of their
corresponding bounds excludes finite first-order stationary points whose
thresholded graph is cyclic.
\end{proposition}

\begin{proof}
Let $Q_C=\prod_{e\in C_L}g_e$, $H^{(k)}=\exp(U^{(k)}/\tau_c)$ entrywise and $T=H^{(0)}$, so $T_{ij}=g_{ij}^{1/\tau_c}$ on allowed edges and zero elsewhere. The soft recursion is $H^{(k+1)}=H^{(k)}+(H^{(k)})^2$ with ordinary matrix products, hence $H^{(K)}=\sum_{\ell=1}^{b}c_{K,\ell}T^\ell$ with integer coefficients $c_{K,\ell}\ge1$ that do not depend on the gates. On an isolated simple cycle only powers $\ell=mL$ reach a diagonal entry, $(T^{mL})_{vv}=Q_C^{m/\tau_c}$, so for every $v\in C_L$
\begin{equation}
u:=U_{vv}^{(K)}
=\tau_c\log\sum_{m=1}^{\lfloor b/L\rfloor}a_m Q_C^{m/\tau_c},
\qquad a_m=c_{K,mL}.
\label{eq:cycle-pressure-walk-expansion}
\end{equation}
Since one complete traversal is feasible and $a_1\ge1$, $u\ge\log Q_C\ge L\log\delta$, and $r:=\partial u/\partial\log Q_C=\sum_m m\omega_m\ge1$ with $\omega_m\propto a_mQ_C^{m/\tau_c}$ summing to one; the bound on $r$ comes from differentiating Eq.~\eqref{eq:cycle-pressure-walk-expansion}, since every closed walk traverses the designated edge at least once, and not from a score inequality. Because $\partial\log Q_C/\partial z_q=(1-q)/\tau_g$ and the cycle contributes $L$ diagonal losses,
\[
\begin{aligned}
\frac{\partial\mathcal L_{\mathrm{raw}}}{\partial z_q}
&=\frac{L(1-q)}{d\tau_g}r\sigma(u)
 \ge\frac{L(1-q)}{d\tau_g}\sigma(L\log\delta),\\
\frac{\partial\mathcal L_{\mathrm{SMNC}}}{\partial z_q}
&=\frac{L(1-q)}{db\tau_g}r\sigma(u/b)
 \ge\frac{L(1-q)}{db\tau_g}\sigma((L/b)\log\delta),
\end{aligned}
\]
and other disconnected blocks contribute nothing to $z_q$. The fidelity derivative is $-\lambda_{\mathrm{fid}}q(q_0-q)(1-q)/\tau_g$ with $q(q_0-q)\le q_0^2/4$. On this cycle the $s=1$ log-det term is $-\log(1-Q_C^2)$, with $\partial h_1/\partial z_q=\frac{1-q}{\tau_g}\frac{2Q_C^2}{1-Q_C^2}\ge0$. Discarding this nonnegative help and using $1/\sigma(x)=1+e^{-x}$, either closure gradient strictly exceeds the opposing fidelity gradient under its bound in Eq.~\eqref{eq:cycle-sufficient-weight-bounds}; for log-det alone, $\nu\,2\delta^{2L}/(1-\delta^{2L})>\lambda_{\mathrm{fid}}q_0^2/4$ gives $B_{\mathrm{ld}}$. At finite logits $(1-q)/\tau_g>0$, so an intact cycle would have a nonzero designated-edge derivative, contradicting first-order stationarity. Applying this to every cycle proves the statement about the maximum bound.
\end{proof}

\paragraph{Length dependence and scope.}
For fixed $\delta,q_0,\lambda_{\mathrm{fid}}$ and a family with $d=\Theta(L)$, so that $b=2^{\lceil\log_2d\rceil}=\Theta(d)$ and $L/b=\Theta(1)$, the sufficient expressions obey $B_{\mathrm{raw}}=\Theta(\delta^{-L})$, $B_{\mathrm{SMNC}}=\Theta(L)$ and $B_{\mathrm{ld}}=\Theta(\delta^{-2L})$. They are sufficient per-cycle bounds, not minimal weights, and they are not uniform over heterogeneous cycle families: a fixed-length short cycle with growing $d$ has an SMNC bound that scales as $\Theta(d^2)$. The closure bounds ignore the common log-det gradient, and $B_{\mathrm{ld}}$ concerns an adjustable multiplier $\nu$ that is fixed at one in the reported ablation.

For the two-cycle probe ($d=b=64$, $q_0=0.2$, $\lambda_{\mathrm{fid}}=5$, $\delta=0.1$), $B_{\mathrm{SMNC}}(L;64)=\frac{204.8}{L}(1+10^{L/64})$, so $B_{\mathrm{SMNC}}(8;64)\approx59.74$, and the bounds for $L=24,32,40,48$ are approximately $28.77,26.64,26.71,28.26$. The proposition rules out cancellation on an intact isolated cycle. It does not guarantee success within a fixed number of updates or that only a minimum-cost edge is removed, and it does not cover a trainable magnitude branch, general data-fitting losses, clipped logits, or overlapping cycles whose closed walks can avoid the designated edge.

\newpage

\section{Additional synthetic results}
\label{app:more-results}

\subsection[Across-d overview and compact tables]{Across-\(d\) overview and compact tables}
\label{app:overview}

\begin{figure}[H]
  \centering
  \includegraphics[width=\linewidth]{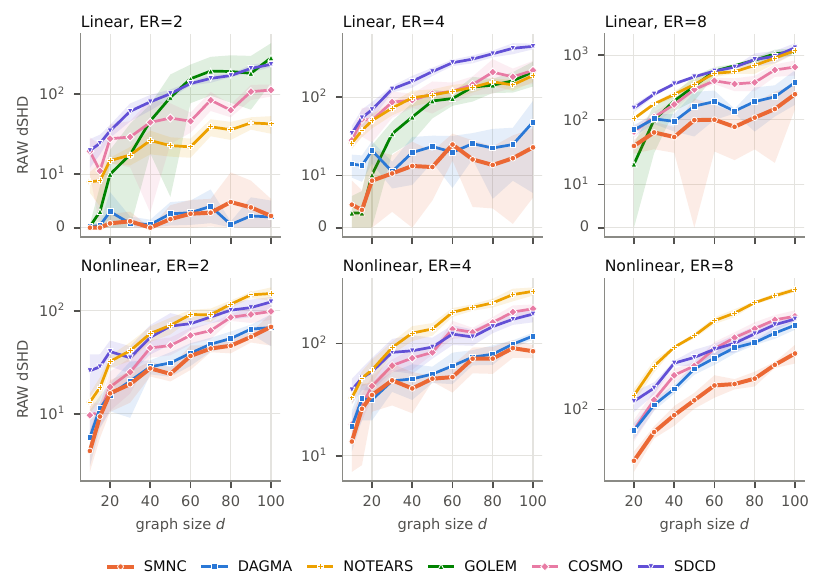}
  \caption{RAW dSHD across graph size \(d\) for ER\(\,=2,4,8\) (columns), Gaussian linear (top) and nonlinear (bottom) SEMs. Lines show the mean over 5 datasets and bands one standard deviation; lower is better. SMNC denotes bounded WG + SMNC (linear) and induced + SMNC (nonlinear). GOLEM was not run on the nonlinear SEMs.}
  \label{fig:overview_raw_dshd}
\end{figure}

Figure~\ref{fig:overview_raw_dshd} shows RAW dSHD across graph sizes for ER\(\,\in\{2,4,8\}\) on linear and nonlinear SEMs, for every method we ran. Tables~\ref{tab:compact_raw_dshd}--\ref{tab:compact_topk_dshd} list all methods at \(d=100\) for every edge density. TOPK keeps up to \(s_0\) edges and is an oracle-sparsity diagnostic. Results for the other graph sizes and for the remaining metrics come from the same archived runs and are omitted for length.

\begin{table}[H]
\centering
\scriptsize
\setlength{\tabcolsep}{3pt}
\caption{RAW dSHD at \(d=100\) (mean \(\pm\) standard deviation over 5 datasets; lower is better). SMNC is bounded WG + SMNC (linear) and induced + SMNC (nonlinear); bold marks the best method in a row. NOTEARS is the DAGrad Adam implementation with the exponential penalty, untuned (Appendix~\ref{app:baseline-protocol}). GOLEM was not run on the nonlinear SEMs (--). GOLEM's RAW entries report its own post-hoc pruned output (Appendix~\ref{app:baseline-protocol}).}
\label{tab:compact_raw_dshd}
\resizebox{\linewidth}{!}{%
\begin{tabular}{llccccccc}
\toprule
SEM & ER & \(s_0\) & SMNC & DAGMA & NOTEARS & GOLEM & COSMO & SDCD \\
\midrule
Linear & 2 & 200 & 2.20 $\pm$ 3.19 & \textbf{2.00 $\pm$ 3.08} & 42.60 $\pm$ 10.53 & 285.00 $\pm$ 148.04 & 112.80 $\pm$ 19.46 & 236.20 $\pm$ 17.84 \\
 & 4 & 400 & \textbf{22.80 $\pm$ 17.05} & 48.00 $\pm$ 41.36 & 188.60 $\pm$ 32.79 & 209.80 $\pm$ 73.46 & 218.80 $\pm$ 71.45 & 444.00 $\pm$ 37.25 \\
 & 5 & 500 & \textbf{16.80 $\pm$ 6.30} & 42.20 $\pm$ 16.86 & 247.60 $\pm$ 21.96 & 261.20 $\pm$ 81.94 & 259.20 $\pm$ 64.40 & 545.00 $\pm$ 39.52 \\
 & 6 & 600 & \textbf{20.80 $\pm$ 15.01} & 88.20 $\pm$ 47.33 & 411.00 $\pm$ 83.81 & 529.40 $\pm$ 173.66 & 324.40 $\pm$ 77.75 & 682.60 $\pm$ 64.06 \\
 & 7 & 700 & \textbf{58.80 $\pm$ 22.17} & 102.20 $\pm$ 47.01 & 637.40 $\pm$ 92.05 & 944.20 $\pm$ 72.19 & 424.00 $\pm$ 89.53 & 849.40 $\pm$ 32.16 \\
 & 8 & 800 & \textbf{248.40 $\pm$ 103.58} & 376.80 $\pm$ 206.54 & 1170.80 $\pm$ 130.44 & 1136.80 $\pm$ 133.16 & 653.40 $\pm$ 170.38 & 1296.40 $\pm$ 227.89 \\
\midrule
Nonlinear & 2 & 200 & 69.80 $\pm$ 23.86 & \textbf{68.80 $\pm$ 23.53} & 147.00 $\pm$ 21.59 & -- & 98.60 $\pm$ 17.04 & 121.80 $\pm$ 14.46 \\
 & 4 & 400 & \textbf{85.40 $\pm$ 11.28} & 116.60 $\pm$ 8.08 & 291.80 $\pm$ 24.47 & -- & 204.20 $\pm$ 18.49 & 183.00 $\pm$ 28.47 \\
 & 5 & 500 & \textbf{116.40 $\pm$ 31.17} & 160.40 $\pm$ 34.62 & 371.00 $\pm$ 41.52 & -- & 269.00 $\pm$ 42.53 & 222.00 $\pm$ 15.08 \\
 & 6 & 600 & \textbf{127.00 $\pm$ 23.40} & 204.20 $\pm$ 6.02 & 486.40 $\pm$ 20.28 & -- & 324.00 $\pm$ 26.84 & 264.00 $\pm$ 32.07 \\
 & 7 & 700 & \textbf{173.60 $\pm$ 28.08} & 290.60 $\pm$ 12.66 & 609.00 $\pm$ 18.28 & -- & 412.40 $\pm$ 31.39 & 335.80 $\pm$ 16.59 \\
 & 8 & 800 & \textbf{247.00 $\pm$ 36.86} & 390.40 $\pm$ 27.61 & 692.60 $\pm$ 26.41 & -- & 450.80 $\pm$ 45.69 & 431.00 $\pm$ 15.89 \\
\bottomrule
\end{tabular}}
\end{table}

\begin{table}[H]
\centering
\scriptsize
\setlength{\tabcolsep}{3pt}
\caption{RAW F1-dir at \(d=100\) (mean \(\pm\) standard deviation over 5 datasets; higher is better). Conventions as in Table~\ref{tab:compact_raw_dshd}.}
\label{tab:compact_raw_f1}
\resizebox{\linewidth}{!}{%
\begin{tabular}{llccccccc}
\toprule
SEM & ER & \(s_0\) & SMNC & DAGMA & NOTEARS & GOLEM & COSMO & SDCD \\
\midrule
Linear & 2 & 200 & \textbf{0.99 $\pm$ 0.01} & \textbf{0.99 $\pm$ 0.01} & 0.88 $\pm$ 0.03 & 0.58 $\pm$ 0.15 & 0.74 $\pm$ 0.04 & 0.56 $\pm$ 0.02 \\
 & 4 & 400 & \textbf{0.97 $\pm$ 0.02} & 0.94 $\pm$ 0.05 & 0.72 $\pm$ 0.05 & 0.78 $\pm$ 0.06 & 0.75 $\pm$ 0.07 & 0.47 $\pm$ 0.05 \\
 & 5 & 500 & \textbf{0.98 $\pm$ 0.01} & 0.96 $\pm$ 0.02 & 0.70 $\pm$ 0.02 & 0.78 $\pm$ 0.06 & 0.76 $\pm$ 0.05 & 0.43 $\pm$ 0.04 \\
 & 6 & 600 & \textbf{0.98 $\pm$ 0.01} & 0.93 $\pm$ 0.04 & 0.60 $\pm$ 0.07 & 0.66 $\pm$ 0.09 & 0.76 $\pm$ 0.05 & 0.29 $\pm$ 0.10 \\
 & 7 & 700 & \textbf{0.96 $\pm$ 0.01} & 0.93 $\pm$ 0.03 & 0.51 $\pm$ 0.05 & 0.49 $\pm$ 0.05 & 0.73 $\pm$ 0.05 & 0.09 $\pm$ 0.04 \\
 & 8 & 800 & \textbf{0.84 $\pm$ 0.06} & 0.78 $\pm$ 0.11 & 0.32 $\pm$ 0.05 & 0.21 $\pm$ 0.06 & 0.65 $\pm$ 0.08 & 0.02 $\pm$ 0.01 \\
\midrule
Nonlinear & 2 & 200 & \textbf{0.83 $\pm$ 0.05} & 0.82 $\pm$ 0.06 & 0.50 $\pm$ 0.09 & -- & 0.69 $\pm$ 0.06 & 0.71 $\pm$ 0.03 \\
 & 4 & 400 & \textbf{0.89 $\pm$ 0.01} & 0.84 $\pm$ 0.01 & 0.45 $\pm$ 0.07 & -- & 0.67 $\pm$ 0.05 & 0.76 $\pm$ 0.04 \\
 & 5 & 500 & \textbf{0.88 $\pm$ 0.03} & 0.82 $\pm$ 0.04 & 0.43 $\pm$ 0.11 & -- & 0.64 $\pm$ 0.09 & 0.76 $\pm$ 0.01 \\
 & 6 & 600 & \textbf{0.89 $\pm$ 0.02} & 0.80 $\pm$ 0.01 & 0.34 $\pm$ 0.04 & -- & 0.64 $\pm$ 0.04 & 0.75 $\pm$ 0.04 \\
 & 7 & 700 & \textbf{0.87 $\pm$ 0.02} & 0.75 $\pm$ 0.02 & 0.25 $\pm$ 0.04 & -- & 0.59 $\pm$ 0.04 & 0.71 $\pm$ 0.02 \\
 & 8 & 800 & \textbf{0.82 $\pm$ 0.03} & 0.69 $\pm$ 0.03 & 0.24 $\pm$ 0.05 & -- & 0.62 $\pm$ 0.06 & 0.66 $\pm$ 0.02 \\
\bottomrule
\end{tabular}}
\end{table}

\begin{table}[H]
\centering
\scriptsize
\setlength{\tabcolsep}{3pt}
\caption{TOPK dSHD at \(d=100\) (mean \(\pm\) standard deviation over 5 datasets; lower is better). TOPK keeps up to \(s_0\) edges and is an oracle-sparsity diagnostic. Conventions as in Table~\ref{tab:compact_raw_dshd}.}
\label{tab:compact_topk_dshd}
\resizebox{\linewidth}{!}{%
\begin{tabular}{llccccccc}
\toprule
SEM & ER & \(s_0\) & SMNC & DAGMA & NOTEARS & GOLEM & COSMO & SDCD \\
\midrule
Linear & 2 & 200 & \textbf{1.60 $\pm$ 2.19} & 2.00 $\pm$ 3.08 & 48.80 $\pm$ 13.01 & 94.00 $\pm$ 38.03 & 96.00 $\pm$ 17.32 & 155.20 $\pm$ 15.53 \\
 & 4 & 400 & \textbf{20.00 $\pm$ 14.35} & 51.60 $\pm$ 42.76 & 227.60 $\pm$ 44.91 & 110.00 $\pm$ 30.17 & 183.20 $\pm$ 48.67 & 424.00 $\pm$ 44.07 \\
 & 5 & 500 & \textbf{19.20 $\pm$ 10.16} & 43.40 $\pm$ 19.36 & 316.00 $\pm$ 25.69 & 153.60 $\pm$ 42.27 & 221.60 $\pm$ 60.60 & 580.00 $\pm$ 35.97 \\
 & 6 & 600 & \textbf{21.20 $\pm$ 16.41} & 88.40 $\pm$ 46.40 & 498.40 $\pm$ 73.59 & 299.20 $\pm$ 114.88 & 280.40 $\pm$ 57.70 & 882.80 $\pm$ 112.81 \\
 & 7 & 700 & \textbf{49.20 $\pm$ 10.26} & 101.80 $\pm$ 46.14 & 686.80 $\pm$ 72.30 & 656.80 $\pm$ 102.22 & 370.80 $\pm$ 66.28 & 1280.00 $\pm$ 36.33 \\
 & 8 & 800 & \textbf{246.00 $\pm$ 99.12} & 348.60 $\pm$ 174.32 & 1109.60 $\pm$ 98.00 & 1231.20 $\pm$ 63.70 & 551.60 $\pm$ 116.66 & 1553.20 $\pm$ 20.43 \\
\midrule
Nonlinear & 2 & 200 & \textbf{61.00 $\pm$ 20.40} & 82.80 $\pm$ 21.98 & 208.40 $\pm$ 23.00 & -- & 114.80 $\pm$ 22.21 & 102.00 $\pm$ 11.58 \\
 & 4 & 400 & \textbf{85.60 $\pm$ 11.63} & 137.60 $\pm$ 5.73 & 370.80 $\pm$ 54.20 & -- & 196.80 $\pm$ 18.47 & 202.40 $\pm$ 23.38 \\
 & 5 & 500 & \textbf{116.40 $\pm$ 31.17} & 192.00 $\pm$ 55.71 & 442.00 $\pm$ 84.30 & -- & 250.80 $\pm$ 29.69 & 276.40 $\pm$ 13.52 \\
 & 6 & 600 & \textbf{127.00 $\pm$ 23.40} & 229.60 $\pm$ 35.76 & 519.20 $\pm$ 32.91 & -- & 291.20 $\pm$ 31.58 & 393.60 $\pm$ 53.56 \\
 & 7 & 700 & \textbf{173.60 $\pm$ 28.08} & 292.00 $\pm$ 17.44 & 650.00 $\pm$ 25.02 & -- & 371.60 $\pm$ 21.65 & 502.40 $\pm$ 44.57 \\
 & 8 & 800 & \textbf{247.00 $\pm$ 36.86} & 369.60 $\pm$ 14.79 & 692.00 $\pm$ 81.30 & -- & 430.00 $\pm$ 24.33 & 665.60 $\pm$ 50.43 \\
\bottomrule
\end{tabular}}
\end{table}

\subsection{Baseline protocol}
\label{app:baseline-protocol}

Baseline optimization budgets, numerical precision, and hyperparameters were
not matched to our own method's protocol. We disclose the re-verified
configuration behind the linear DAGMA, GOLEM, COSMO and NOTEARS numbers and the
nonlinear DAGMA and NOTEARS numbers in Table~\ref{tab:baseline_protocol_disclosure}.
Every entry below was re-verified against the configuration recorded with each
method's own run outputs and against the defaults of the benchmark driver
that invoked the method. The table notes give the sources.
The SDCD configuration was not archived. The SDCD numbers are consistent
with SDCD's default settings applied to the unstandardized benchmark data.
Such a run would be outside SDCD's documented protocol, which z-scores each
variable before fitting.

\paragraph{Provenance of the SMNC numbers.}
The SMNC entries in Table~\ref{tab:main_raw_d100_summary} and in
Appendix~\ref{app:overview}
come from reruns of SMNC on the benchmark datasets. We archived the configuration, the
dataset file and the seed of every rerun, so each reported cell can be recomputed from its
\texttt{metrics.json}. They replace the values in the reviewed version, which came from runs
whose configuration files were not archived. The reviewed and rerun values agree
to within one reported standard deviation in every dSHD cell, RAW and TOPK, and the
differences have no consistent sign.

For the linear SEMs, the archived configuration sets some values per dataset. The optimizer is Adam with learning rate \(5\times10^{-3}\) and betas \((0.99,0.999)\); the magnitude bound is \(\beta=5\), and training runs \(T=5\) stages with \(s=1.0,0.9,0.8,0.7,0.6\). On 283 of the 305 datasets it uses \(\lambda_1=10^{-2}\). It uses \(\lambda_1=1.2\times10^{-2}\) on 2 datasets and \(1.25\times10^{-2}\) on 15, all with \(d\in\{60,\dots,90\}\) and ER\(\in\{6,8\}\), and \(\lambda_1=3\times10^{-2}\) on the five \(d=100\), ER\(=8\) datasets. The magnitude threshold is \(0.1\) on 300 datasets and \(0.3\) on those five. The first four stages run \(2\times10^{4}\) steps each when \(d\ge40\) and ER\(\ge5\) (140 datasets), and \(10^{4}\) steps otherwise (165 datasets). The last stage always runs \(10^{4}\) steps. The cycle weight \(\mu_{\mathrm{cycle}}\) is multiplied by \(0.1\) after each stage, and the cycle term is applied at every second gradient step. The configuration sets \((\mu_{\mathrm{init}},\mu_{\mathrm{cycle}})=(10,0.1)\) on 289 datasets, and \((0.3,0.3)\) on the 16 datasets whose sample covariance has a condition number of at least \(3.2\times10^{8}\). No dataset lies on that cut: the two nearest lie at \(3.08\times10^{8}\) and \(3.14\times10^{8}\) below it and at \(3.31\times10^{8}\) and \(3.46\times10^{8}\) above it. The tolerance of the barrier's inverse-nonnegativity check is 0. For the nonlinear SEMs, it uses Adam with learning rate \(2\times10^{-4}\), hidden width 20, \(\lambda_2=5\times10^{-3}\), threshold \(0.3\), \(T=5\) stages of \(10^{4}\) steps each, \(s=1\), and \((\mu_{\mathrm{init}},\mu_{\mathrm{cycle}})=(0.3,0.1)\), with \(\mu_{\mathrm{cycle}}\) multiplied by \(0.1\) after each stage and the cycle term applied at every step. It uses \(\lambda_1=1.25\times10^{-2}\) on the five \(d=100\), ER\(=5\) datasets and \(\lambda_1=10^{-2}\) on the other 300. The archive records these values for every dataset but not how they were chosen: how the per-dataset \(\lambda_1\) and threshold values and the condition-number switch were selected is not recorded.

\paragraph{Compute resources.}
NOTEARS was run through DAGrad on an NVIDIA A30 GPU; at \(d=100\) its median fit time was 29.9\,s for linear and 1332\,s for nonlinear SEMs. DAGMA's median fit time at \(d=100\) was 42.6\,s (linear) and 705.7\,s (nonlinear); the GPU model was not recorded with these times. The SMNC runs on the Sachs data used an NVIDIA A30 with PyTorch 2.4.0, and their run time was not recorded. The hardware, run time and memory of the synthetic SMNC runs were not recorded, and neither were the total compute and the compute of preliminary or failed runs.

\paragraph{Licences.}
DAGMA and NOTEARS are released under the Apache-2.0 licence, and SDCD under the MIT licence. DAGrad, through which the NOTEARS numbers were produced, has a non-commercial licence. GOLEM's upstream code is Apache-2.0, but we did not confirm the licence of the copy we ran, and we did not verify a licence for the COSMO code. The Sachs data~\cite{sachs} come with no stated licence; our runs read a laboratory CSV copy of them, and the 17-edge reference DAG is the one distributed with bnlearn (Section~\ref{sec:sachs}).

\paragraph{Scope of the attribution.}
No benchmark in this paper isolates scale normalization from the closure head and the factorization: raw closure is compared with SMNC only in the controlled probes of Appendix~\ref{app:controlled-long-cycle}, and the benchmark comparison with DAGMA also differs in iteration budget, hyperparameters and, for nonlinear SEMs, network width (Table~\ref{tab:baseline_protocol_disclosure}).

We also ran DAGMA on the 45 linear ER\(=8\) datasets
(\(d\in\{20,30,\dots,100\}\)) with the archived SMNC settings. These cover
\(\lambda_1\), the threshold, the learning rate, the step counts and the
\(\mu\) and \(s\) schedules. DAGMA's mean RAW dSHD
was \(215.3\), against \(172.5\) with the settings in
Table~\ref{tab:baseline_protocol_disclosure} and \(104.2\) for the SMNC rerun.
The SMNC rerun had a lower dSHD than the matched DAGMA run on 37 of the 45
datasets.

A trial point is outside the domain of \(h_s\) in Eq.~\eqref{eq:dagma-barrier} when \(s\le 0\) or \(\rho(A\circ A)\ge s\). A positive determinant does not certify the domain: \(sI-A\circ A\) must be a nonsingular M-matrix.
Linear DAGMA, as run from the authors' solver, rejects a step when the inverse of \(sI-A\circ A\), offset by \(10^{-16}\), has a negative entry, which tests the M-matrix domain. Nonlinear DAGMA evaluates \(-\log|\det(sI-A\circ A)|+d\log s\) and rejects a step only when that value is negative, and our nonlinear SMNC solver rejects a step only when the value is non-finite or negative. Both can therefore accept points outside the domain: a point with a negative determinant, or with a positive determinant but a nonpositive principal minor, gives a finite value that is not the barrier above. Our linear SMNC solver uses the inverse test, with tolerance 0 in the archived runs.
GOLEM's acyclicity penalty is finite for every real adjacency, so it does not reject steps for leaving this domain.
COSMO parameterizes an acyclic orientation from node priorities, so every parameter value stays in that class and there is no log-det domain to leave.
SDCD constrains a spectral radius rather than \(h_s\), so it has no log-det domain and does not reject steps on these grounds.

\begin{table}[p]
\centering
\caption{Verified optimization protocol for the benchmark baselines. ``Own-code default'' is the budget or
behavior the method's own vendored script uses when no override is passed.
``Used'' is the value actually recorded with the runs
behind the reported numbers. SMNC's own protocol is stated in
Sections~\ref{sec:synth} and~\ref{sec:metrics}, and its archived per-dataset
settings are given in this appendix.}
\label{tab:baseline_protocol_disclosure}
\footnotesize
\setlength{\tabcolsep}{3pt}
\begin{tabular}{@{}p{0.5in}p{1.05in}p{0.55in}p{0.42in}p{2.6in}@{}}
\toprule
Method & Steps used (vs.\ own default) & Precision & RAW col.\ & Notes \& hyperparameters \\
\midrule
DAGMA (linear) &
90{,}000 (4\(\times\)20k warm + 10k final) vs.\ 180{,}000 (4\(\times\)30k + 60k) &
float64 &
thresh.\ &
\(w_{\mathrm{threshold}}=0.3\); \(\lambda_1{=}0.03\), lr\(=3{\times}10^{-4}\)
are the vendored script's own defaults, unmodified; only the iteration
budget was reduced.\textsuperscript{a} \\[6pt]
DAGMA (nonlinear) &
40{,}000 (3\(\times\)10k warm + 10k final) vs.\ 230{,}000 (3\(\times\)50k + 80k) &
float64 &
thresh.\ &
MLP with one hidden layer of width 10, as in DAGMA's own nonlinear example;
\(\lambda_1=0.02\), \(\lambda_2=0.005\), lr\(=2{\times}10^{-4}\), \(T=4\),
\(\mu_{\mathrm{init}}=0.1\), \(\mu\) factor \(0.1\), \(s=1\),
\(w_{\mathrm{threshold}}=0.3\), all equal to the defaults of DAGMA's nonlinear
\texttt{fit}; only the iteration budget was reduced.\textsuperscript{d} \\[6pt]
NOTEARS (DAGrad) &
5\(\times\)1{,}000 Adam steps vs.\ up to 100 outer L-BFGS-B solves in NOTEARS's own code &
float64 &
thresh.\ &
DAGrad's PyTorch implementation of the NOTEARS objective with the exponential penalty
\(\operatorname{tr}e^{A\circ A}-d\) and Adam, \textbf{not} the original SciPy L-BFGS-B solver.
Linear: \(\ell_2\) loss, \(\lambda_1=0.1\), lr\(=10^{-3}\), \(h_{\mathrm{tol}}=10^{-8}\),
\(w_{\mathrm{threshold}}=0.3\). Nonlinear: a one-hidden-layer sigmoid MLP of width 40,
\(\lambda_2=0.01\), same budget and threshold. These are the runner's defaults, the same for every
dataset.\textsuperscript{e} \\[6pt]
GOLEM &
50{,}000 vs.\ 100{,}000 (half the script's default) &
float32 (script default) &
\textbf{proc.} &
\textbf{RAW columns report GOLEM's post-hoc pruned (``processed'')
output.} Its raw output is dense (e.g.\ directed-edgewise SHD
\(9500\) for one \(d{=}100\), ER\(=\)4 run, vs.\ \(209.80\pm73.46\), the
table's processed-output mean over 5 datasets at that setting).
\(\lambda_1=0.02\), \(\lambda_2=5.0\) are the script's own equal-variance
defaults, unmodified; only the iteration budget was reduced.\textsuperscript{b} \\[6pt]
COSMO &
2{,}000 epochs, batch 64, at one fixed config; n/a own default (grid-search method) &
float64 &
thresh.\ &
\textbf{evaluated at one fixed hyperparameter point}: the geometric
midpoint of each parameter's published search range
(lr\(=3.16{\times}10^{-3}\), \(\ell_1\)-reg\(=3.16{\times}10^{-4}\),
\(\ell_2\)-reg\(=2.24{\times}10^{-3}\), priority-reg\(=1.73{\times}10^{-3}\),
temperature\(=7.07{\times}10^{-4}\), shift\(=10^{-2}\)), with
threshold\(=0.3\). COSMO's own documented protocol runs a grid search per
dataset and keeps the configuration with the best validation
ROC-AUC.\textsuperscript{c} \\
\bottomrule
\end{tabular}

\vspace{4pt}
\raggedright
\footnotesize
\textsuperscript{a}Steps used are the benchmark driver's defaults
(\texttt{warm\_iter}\(=20000\), \texttt{max\_iter}\(=10000\), \(T=5\)) passed to the
DAGMA authors' \texttt{DagmaLinear.fit}, whose own documented defaults are
\texttt{warm\_iter}\(=3\times10^4\), \texttt{max\_iter}\(=6\times10^4\); the
outer loop runs \((T{-}1)\cdot\texttt{warm\_iter}+\texttt{max\_iter}\) steps.\\
\textsuperscript{b}GOLEM's run script defaults to \(10^5\) iterations in
\texttt{float32} with equal-variance \(\lambda_1=2\times10^{-2}\),
\(\lambda_2=5.0\); every archived run records \texttt{float32} and \(5\times10^4\)
iterations. GOLEM records both its raw and its processed output; the RAW
columns use the processed one.\\
\textsuperscript{c}COSMO's benchmark driver hardcodes the six values listed
above, each the geometric midpoint of the log-uniform search range published
with COSMO's code; COSMO's documented grid-search-and-validate invocation is
not called. \texttt{threshold}, \texttt{batch\_size}, and \texttt{max\_epochs}
are fixed generic parameters in both COSMO's grid file and its default
configuration.\\
\textsuperscript{d}All 305 archived nonlinear DAGMA runs (61 settings with 5 datasets each) record
the same configuration (hidden width 10, \texttt{warm\_iter}\(=\)\texttt{max\_iter}\(=10^4\),
\(T=4\)); DAGMA's \texttt{DagmaNonlinear.fit} defaults to
\texttt{warm\_iter}\(=5\times10^4\) and \texttt{max\_iter}\(=8\times10^4\), and its outer
loop runs \((T{-}1)\cdot\texttt{warm\_iter}+\texttt{max\_iter}\) steps. The MLP
width has no default in DAGMA's code; width 10 is the value used in DAGMA's
own example. The numerical precision is not recorded with these runs.\\
\textsuperscript{e}All 305 archived linear and 305 archived nonlinear DAGrad runs (61 settings with 5 datasets
each) record this configuration and ran on an NVIDIA A30. On 5 linear runs, at
\((d,\mathrm{ER})\in\{(40,2),(40,8),(50,5),(70,2),(90,2)\}\), the returned \(h\) exceeds the tolerance
\(10^{-8}\) and the RAW output is not a DAG; these runs are scored as returned. The runs record neither the
DAGrad nor the NOTEARS code revision. NOTEARS's TOPK column is the oracle edge-count projection of its raw
weights, a diagnostic and not the baseline's standard output.
\end{table}


\subsection{Sachs reachability-cue construction}
\label{app:sachs_order_cues}

\paragraph{Training protocol and reproduction.}
The reported runs use the pooled, column-wise z-scored Sachs data, seed 0,
four outer stages, initial \(\mu_{\mathrm{init}}=0.1\) with factor 0.1,
and output threshold \(0.3\). The cycle-enabled variants use initial
\(\mu_{\mathrm{cycle}}=0.3\) with factor 0.1; the active soft losses use
initial \(\mu_{\mathrm{rel}}=0.5\) with factor 0.99. The cycle-off row uses
the same internal nonlinear backbone with \(\mu_{\mathrm{cycle}}=0\).
Archived CPU/GPU reruns reproduce the thresholded directed graphs of all
four cycle-enabled 12-cue/no-cue variants. These are fixed-seed
reproductions, not additional independent-seed trials; the cycle-off row
was not included in that reproduction study.

\paragraph{Reference network.}
We use the standard 17-edge Sachs reference DAG \(\mathcal G_{17}\) distributed with \texttt{bnlearn} (\url{https://www.bnlearn.com/research/sachs05/}) on the eleven variables Raf, Mek, Plcg, PIP2, PIP3, Erk, Akt, PKA, PKC, P38, Jnk. Its edges are PKC\(\to\){PKA, Raf, Mek, P38, Jnk}; PKA\(\to\){Raf, Mek, Erk, Akt, P38, Jnk}; Raf\(\to\)Mek; Mek\(\to\)Erk; Erk\(\to\)Akt; Plcg\(\to\){PIP3, PIP2}; PIP3\(\to\)PIP2.

\paragraph{Cue sources.}
A cue \(i\leadsto j\) means that \(i\) is upstream of \(j\) in the closure sense, possibly through a multi-hop path. Cues come from matched intervention/reference comparisons in the continuous Sachs conditions: Mek (\texttt{u0126}), Akt (\texttt{aktinhib}), PKC (\texttt{g0076}) and PIP2 (\texttt{psitect}) interventions, each against its matching \texttt{cd3cd28} or \texttt{cd3cd28\_icam2} background, and PKA activation (\texttt{b2camp.csv}) against both backgrounds. The PKA rows use a PKA-activation condition and not a stimulus-matched background, so we treat that evidence as weaker. For each target \(i\) and response \(j\neq i\) we compare \(X_j\) under the intervention and under the matched reference by the Kolmogorov--Smirnov statistic \(D_{ij}\), the robust median effect \(E_{ij}=(\operatorname{median}X_j^{\mathrm{int}}-\operatorname{median}X_j^{\mathrm{ref}})/(\operatorname{IQR}X_j^{\mathrm{ref}}+10^{-8})\), and \(S^{\mathrm{stat}}_{ij}=D_{ij}|E_{ij}|\). Pairs are retained when \(D_{ij}\ge0.20\), \(|E_{ij}|\ge0.30\) and \(S^{\mathrm{stat}}_{ij}\ge0.20\), and pairs seen in several backgrounds are ranked by \(\max_bS^{\mathrm{stat},(b)}_{ij}+0.25\operatorname{mean}_bS^{\mathrm{stat},(b)}_{ij}+0.20\,n_{\mathrm{bg}}(i,j)\), where \(n_{\mathrm{bg}}\) is the number of supporting backgrounds.

\paragraph{Reference-filtered 12-cue set.}
This step is oracle-assisted: we use the reference closure to filter unstable directions. Akt and PIP2 are excluded as cue sources, because they produced more no-path or reverse-direction candidates in preliminary screening; they may still be responses. The set takes the seven highest-scoring PKC-sourced pairs, the two Mek-sourced pairs whose responses are downstream of Mek in the reference (Erk, Akt), and the three highest-scoring PKA-sourced pairs from \texttt{b2camp.csv} (PKC is excluded as a response because it is upstream of PKA in \(\mathcal G_{17}\)); responses are restricted to the eight protein-kinase nodes (excluding Plcg, PIP2, PIP3). The 12 relations are PKC\(\leadsto\){Mek, P38, Akt, Raf, Jnk, Erk, PKA}, Mek\(\leadsto\){Akt, Erk} and PKA\(\leadsto\){Mek, Jnk, P38}. All are consistent with the transitive closure of \(\mathcal G_{17}\), so we describe this as a data-induced, reference-filtered weak-cue experiment and not as unsupervised cue extraction.

\paragraph{Encoding as closure-side supervision.}
For the positive-only variant, the active target set consists of the 12
selected forward pairs with target 1, and no hard exclusions are applied.
The hard-only variant uses the same pairs to forbid reverse direct edges
but has no active relation loss. The hard+soft variant applies these
hard exclusions and supervises both the 12 forward targets at 1 and the
12 reverse targets at 0. Let \(M^{\mathrm{cue}}_{ij}\) be the binary
supervision mask and \(O^{\mathrm{cue}}_{ij}\) its binary target.
For either active soft loss, the implementation averages BCE over the
supervised entries:
\[
\mathcal L_{\mathrm{rel}}=
\frac{\sum_{i\ne j}M^{\mathrm{cue}}_{ij}\,
\ell_{\mathrm{BCE}}(S_{ij},O^{\mathrm{cue}}_{ij})}{\sum_{i\ne j}M^{\mathrm{cue}}_{ij}}.
\]
Here \(S_{ij}\) is the SMNC reachability logit. In the 12-cue comparison,
the denominators are 12 for positive-only and 24 for two-sided supervision. Thus,
comparing positive-only with hard+soft changes the hard mask, reverse
targets, and per-target weighting; it is not a strict factorial
comparison with an identical soft objective. None of these cues requires
a positively supervised pair to be a direct edge.
\subsection{Nonlinear reachability-cue ablation}
\label{app:hard-mask}

Table~\ref{tab:soft_cue_nonlinear} compares No cues, Hard-only,
Soft-only, and Hard+Soft in the nonlinear SMNC model.
Cue-assisted variants use oracle ancestry.
For each \(i\leadsto j\), Hard-only forbids the reverse direct
edge \(j\to i\), while Soft-only assigns forward and reverse
closure targets of \(1\) and \(0\) without hard exclusion.
Hard+Soft combines both operations. A positive reachability target
does not require a direct edge \(i\to j\).

We use \(d=100\), ER\(\in\{2,4,5,6,7,8\}\), and five matched
datasets per setting, with unstandardized and standardized inputs.
All variants use \(\mu_{\mathrm{init}}=0.3\), with initial
\(\mu_{\mathrm{rel}}=0.01\) for soft supervision.
The shared output thresholds \(t\in\{0.3,0.5\}\) are applied
to the same trained matrices.
No cues is this ablation's control; the main cue-free benchmark
uses a separate protocol.

Adding soft supervision to hard exclusion yields more consistent
recovery gains at \(t=0.5\). At \(t=0.3\), gains occur in denser
settings, while sparse-setting outcomes are mixed.
These comparisons are descriptive: the supervision weight and
thresholds were explored using synthetic ground truth without
independent validation.

\begin{table}[H]
\centering
\scriptsize
\setlength{\tabcolsep}{1.8pt}
\caption{Nonlinear soft-supervision ablation with oracle ancestry cues at \(d=100\).
Entries report mean \(\pm\) sample standard deviation over five matched datasets.
Both soft variants use two-sided closure targets.
Bold marks the best mean for each metric within a row, including ties.}
\label{tab:soft_cue_nonlinear}
\resizebox{\linewidth}{!}{%
\begin{tabular}{@{}crcccccccc@{}}
\toprule
& & \multicolumn{2}{c}{No cues}
& \multicolumn{2}{c}{Hard-only}
& \multicolumn{2}{c}{Soft-only}
& \multicolumn{2}{c}{Hard+Soft} \\
\cmidrule(lr){3-4}\cmidrule(lr){5-6}\cmidrule(lr){7-8}\cmidrule(lr){9-10}
\(t\) & ER & dSHD \(\downarrow\) & F1-dir \(\uparrow\)
   & dSHD \(\downarrow\) & F1-dir \(\uparrow\)
   & dSHD \(\downarrow\) & F1-dir \(\uparrow\)
   & dSHD \(\downarrow\) & F1-dir \(\uparrow\) \\
\midrule
\multicolumn{10}{l}{\textit{(a) Unstandardized inputs}} \\
\addlinespace[2pt]
\multirow{6}{*}{0.3} & 2 & 65.8 \(\pm\) 22.3 & 0.839 \(\pm\) 0.051 & \textbf{61.8 \(\pm\) 15.8} & \textbf{0.847 \(\pm\) 0.040} & 80.6 \(\pm\) 24.6 & 0.817 \(\pm\) 0.051 & 75.0 \(\pm\) 16.4 & 0.827 \(\pm\) 0.038 \\
 & 4 & 87.6 \(\pm\) 12.0 & 0.889 \(\pm\) 0.014 & \textbf{81.4 \(\pm\) 13.9} & \textbf{0.897 \(\pm\) 0.017} & 88.0 \(\pm\) 11.4 & 0.891 \(\pm\) 0.012 & 84.8 \(\pm\) 16.6 & 0.895 \(\pm\) 0.020 \\
 & 5 & 117.2 \(\pm\) 31.5 & 0.876 \(\pm\) 0.034 & 102.0 \(\pm\) 27.5 & 0.892 \(\pm\) 0.029 & 116.2 \(\pm\) 31.2 & 0.879 \(\pm\) 0.033 & \textbf{100.2 \(\pm\) 33.1} & \textbf{0.896 \(\pm\) 0.035} \\
 & 6 & 127.6 \(\pm\) 21.0 & 0.886 \(\pm\) 0.020 & 115.8 \(\pm\) 24.7 & 0.897 \(\pm\) 0.024 & 128.0 \(\pm\) 20.6 & 0.888 \(\pm\) 0.020 & \textbf{113.6 \(\pm\) 23.7} & \textbf{0.901 \(\pm\) 0.022} \\
 & 7 & 179.6 \(\pm\) 28.7 & 0.860 \(\pm\) 0.024 & 153.6 \(\pm\) 31.7 & 0.880 \(\pm\) 0.027 & 178.4 \(\pm\) 22.5 & 0.865 \(\pm\) 0.019 & \textbf{146.4 \(\pm\) 27.3} & \textbf{0.888 \(\pm\) 0.023} \\
 & 8 & 244.2 \(\pm\) 37.9 & 0.825 \(\pm\) 0.031 & 216.4 \(\pm\) 30.3 & 0.848 \(\pm\) 0.023 & 229.6 \(\pm\) 35.8 & 0.838 \(\pm\) 0.028 & \textbf{200.8 \(\pm\) 24.6} & \textbf{0.862 \(\pm\) 0.018} \\
\addlinespace[3pt]
\multirow{6}{*}{0.5} & 2 & 47.6 \(\pm\) 18.5 & 0.873 \(\pm\) 0.047 & \textbf{42.2 \(\pm\) 10.2} & 0.887 \(\pm\) 0.029 & 44.6 \(\pm\) 20.5 & 0.886 \(\pm\) 0.051 & \textbf{42.2 \(\pm\) 13.6} & \textbf{0.892 \(\pm\) 0.036} \\
 & 4 & 86.6 \(\pm\) 8.7 & 0.881 \(\pm\) 0.013 & 80.2 \(\pm\) 11.9 & 0.890 \(\pm\) 0.018 & \textbf{76.4 \(\pm\) 4.2} & \textbf{0.897 \(\pm\) 0.007} & 76.8 \(\pm\) 16.0 & 0.896 \(\pm\) 0.024 \\
 & 5 & 143.6 \(\pm\) 14.7 & 0.835 \(\pm\) 0.020 & 130.4 \(\pm\) 12.4 & 0.851 \(\pm\) 0.016 & 138.4 \(\pm\) 20.2 & 0.843 \(\pm\) 0.026 & \textbf{121.2 \(\pm\) 16.8} & \textbf{0.863 \(\pm\) 0.021} \\
 & 6 & 183.2 \(\pm\) 8.4 & 0.823 \(\pm\) 0.010 & 164.6 \(\pm\) 15.2 & 0.843 \(\pm\) 0.017 & 168.0 \(\pm\) 12.1 & 0.840 \(\pm\) 0.013 & \textbf{156.0 \(\pm\) 14.2} & \textbf{0.852 \(\pm\) 0.015} \\
 & 7 & 278.4 \(\pm\) 25.6 & 0.756 \(\pm\) 0.027 & 252.4 \(\pm\) 32.0 & 0.781 \(\pm\) 0.034 & 263.6 \(\pm\) 17.8 & 0.773 \(\pm\) 0.018 & \textbf{237.8 \(\pm\) 34.8} & \textbf{0.796 \(\pm\) 0.036} \\
 & 8 & 420.0 \(\pm\) 75.3 & 0.642 \(\pm\) 0.092 & 385.0 \(\pm\) 56.5 & 0.682 \(\pm\) 0.062 & 403.2 \(\pm\) 77.5 & 0.662 \(\pm\) 0.092 & \textbf{359.6 \(\pm\) 56.1} & \textbf{0.709 \(\pm\) 0.060} \\
\midrule
\multicolumn{10}{l}{\textit{(b) Standardized inputs}} \\
\addlinespace[2pt]
\multirow{6}{*}{0.3} & 2 & 233.4 \(\pm\) 11.3 & 0.442 \(\pm\) 0.034 & \textbf{196.2 \(\pm\) 17.2} & 0.521 \(\pm\) 0.041 & 254.8 \(\pm\) 11.2 & 0.462 \(\pm\) 0.024 & 201.8 \(\pm\) 15.3 & \textbf{0.560 \(\pm\) 0.026} \\
 & 4 & 360.6 \(\pm\) 13.5 & 0.505 \(\pm\) 0.018 & 305.8 \(\pm\) 15.3 & 0.574 \(\pm\) 0.025 & 376.4 \(\pm\) 15.6 & 0.515 \(\pm\) 0.020 & \textbf{301.6 \(\pm\) 15.8} & \textbf{0.607 \(\pm\) 0.017} \\
 & 5 & 440.4 \(\pm\) 40.9 & 0.481 \(\pm\) 0.035 & 360.6 \(\pm\) 30.3 & 0.574 \(\pm\) 0.031 & 433.0 \(\pm\) 54.9 & 0.514 \(\pm\) 0.049 & \textbf{356.8 \(\pm\) 42.4} & \textbf{0.602 \(\pm\) 0.040} \\
 & 6 & 506.4 \(\pm\) 40.8 & 0.482 \(\pm\) 0.040 & 418.0 \(\pm\) 48.3 & 0.570 \(\pm\) 0.055 & 496.4 \(\pm\) 38.9 & 0.523 \(\pm\) 0.038 & \textbf{403.6 \(\pm\) 30.6} & \textbf{0.608 \(\pm\) 0.029} \\
 & 7 & 572.0 \(\pm\) 20.3 & 0.485 \(\pm\) 0.028 & 485.6 \(\pm\) 22.6 & 0.562 \(\pm\) 0.028 & 552.2 \(\pm\) 20.0 & 0.533 \(\pm\) 0.026 & \textbf{454.2 \(\pm\) 22.9} & \textbf{0.615 \(\pm\) 0.025} \\
 & 8 & 651.8 \(\pm\) 49.4 & 0.477 \(\pm\) 0.063 & 547.2 \(\pm\) 32.2 & 0.560 \(\pm\) 0.041 & 650.2 \(\pm\) 39.8 & 0.510 \(\pm\) 0.045 & \textbf{521.4 \(\pm\) 38.1} & \textbf{0.600 \(\pm\) 0.031} \\
\addlinespace[3pt]
\multirow{6}{*}{0.5} & 2 & 209.4 \(\pm\) 8.6 & 0.421 \(\pm\) 0.039 & 177.2 \(\pm\) 14.0 & 0.499 \(\pm\) 0.041 & 211.4 \(\pm\) 8.6 & 0.462 \(\pm\) 0.024 & \textbf{174.6 \(\pm\) 17.3} & \textbf{0.547 \(\pm\) 0.040} \\
 & 4 & 332.2 \(\pm\) 12.9 & 0.468 \(\pm\) 0.016 & 302.2 \(\pm\) 11.6 & 0.520 \(\pm\) 0.024 & 337.2 \(\pm\) 11.5 & 0.478 \(\pm\) 0.016 & \textbf{290.0 \(\pm\) 17.7} & \textbf{0.555 \(\pm\) 0.034} \\
 & 5 & 419.8 \(\pm\) 21.4 & 0.420 \(\pm\) 0.040 & 373.2 \(\pm\) 20.9 & 0.492 \(\pm\) 0.038 & 404.6 \(\pm\) 24.1 & 0.454 \(\pm\) 0.038 & \textbf{361.8 \(\pm\) 22.1} & \textbf{0.525 \(\pm\) 0.033} \\
 & 6 & 518.4 \(\pm\) 28.9 & 0.376 \(\pm\) 0.044 & 466.0 \(\pm\) 26.4 & 0.444 \(\pm\) 0.040 & 499.4 \(\pm\) 26.9 & 0.419 \(\pm\) 0.043 & \textbf{449.8 \(\pm\) 17.1} & \textbf{0.479 \(\pm\) 0.027} \\
 & 7 & 596.0 \(\pm\) 22.7 & 0.363 \(\pm\) 0.034 & 546.0 \(\pm\) 17.1 & 0.424 \(\pm\) 0.026 & 573.2 \(\pm\) 17.3 & 0.410 \(\pm\) 0.028 & \textbf{514.0 \(\pm\) 21.5} & \textbf{0.483 \(\pm\) 0.028} \\
 & 8 & 683.0 \(\pm\) 41.5 & 0.351 \(\pm\) 0.059 & 620.4 \(\pm\) 31.8 & 0.418 \(\pm\) 0.047 & 664.2 \(\pm\) 34.7 & 0.391 \(\pm\) 0.047 & \textbf{605.0 \(\pm\) 32.5} & \textbf{0.444 \(\pm\) 0.046} \\
\bottomrule
\end{tabular}}
\end{table}

\end{document}